\documentclass{article}
\usepackage{graphicx} 
\usepackage{amsmath}
\usepackage[authoryear]{natbib}
\usepackage{float}
\usepackage{textcomp}

\title{Predicting BVD Re-emergence in Irish Cattle From Highly Imbalanced Herd-Level Data Using Machine Learning Algorithms}
\author{Niamh Mimnagh, Andrew Parnell, Conor McAloon, Jaden Carlson, \\ Maria Guelbenzu, Jonas Brock, Damien Barrett,\\ Guy McGrath, Jamie Tratalos, Rafael Moral}
\date{}
\begin{document}

\maketitle

Key-words: targeted testing, anomaly detection, binary classification problem, class imbalance, supervised learning

\section*{Abstract}
Bovine Viral Diarrhoea (BVD) has been the focus of a successful eradication programme in Ireland, with the herd-level prevalence declining from $11.3\%$ in 2013 to just $0.2\%$ in 2023. As the country moves toward BVD freedom, the development of predictive models for targeted surveillance becomes increasingly important to mitigate the risk of disease re-emergence. In this study, we evaluate the performance of a range of machine learning algorithms, including binary classification and anomaly detection techniques, for predicting BVD-positive herds using highly imbalanced herd-level data. We conduct an extensive simulation study to assess model performance across varying sample sizes and class imbalance ratios, incorporating resampling, class weighting, and appropriate evaluation metrics (sensitivity, positive predictive value, F1-score and AUC values). Random forests and XGBoost models consistently outperformed other methods, with the random forest model achieving the highest sensitivity and AUC across scenarios, including real-world prediction of 2023 herd status, correctly identifying 219 of 250 positive herds while halving the number of herds that require compared to a blanket-testing strategy.

\section{Introduction}
Bovine Viral Diarrhoea (BVD) \citep{houe1995epidemiology} is a highly contagious infectious disease, caused by the Bovine Viral Diarrhoea Virus, that can have economically devastating consequences for the dairy and beef industries. In Ireland, it is estimated that, prior to 2013 and the implementation of a mandatory BVD eradication programme, the cost of BVD to Irish farmers totalled approximately €102 million annually \citep{stott2012predicted}. Since the initiation of this eradication programme, the proportion of herds testing positive has decreased from $11.3\%$ of herds in 2013 to $0.39\%$ of breeding herds in 2023 \citep{ahi2024}.

Ireland is moving towards disease freedom, and there is potential to relax individual testing measures currently in place. The goal of this study was to examine the effectiveness of various machine learning methods in predicting re-emergence of BVD in Irish herds.  Such an application would facilitate risk-based surveillance strategies to allow for earlier detection and mitigation of a re-emergence event in a post-eradication scenario. The very low prevalence of the disease is a challenging aspect of the data, and different machine learning strategies need to be used to take this into account.

We detail the implementation of these machine learning methods for classification and anomaly detection, and present their results in predicting the BVD disease-status of herds in Ireland, using data collected between 2013 and 2021. The remainder of this paper is laid out as follows. In Section \ref{ml} we introduce the reader to machine learning and its sub-types. In Section \ref{methods} we describe the machine learning methodologies examined for binary classification, providing a brief overview of their uses, advantages and disadvantages. In Section \ref{imbalanced_data} we present approaches for dealing with highly imbalanced datasets, including the use of appropriate evaluation metrics and resampling techniques. In Section \ref{simulation} we present the results of a simulation study, designed to test the performance of our machine learning methods on data of varying size and  class imbalance. In Section \ref{data} we present the dataset on BVD in Irish cattle herds, and the results obtained predicting BVD occurrence on this dataset.
Finally, in Section~\ref{discussion} we discuss our results in light of the implementation of a surveillance programme in Ireland to prevent BVD re-emergence.

\section{The Use of Machine Learning to Predict BVD Infection}
\label{ml}
Machine learning (ML) techniques have the potential to predict diseases including BVD in Irish cattle herds. By analysing large datasets that include historical health records, herd population factors and details of neighbouring herds, ML algorithms can identify patterns that may not be immediately apparent \citep{neethirajan2020role}. These models may have the potential to assist farmers and veterinarians in making informed decisions relating to the health of Irish cattle, thereby reducing economic losses due to BVD infections.

Various ML techniques may be employed to predict diseases in animal herds \citep{zhang2021application}. Supervised learning models such as decision trees, random forests and support vector machines are commonly used for disease prediction. These models are trained on data where the outcome (whether an Irish cattle herd tested positive for BVD in a given year) is known, and can be used to predict future cases \citep{nasteski2017overview}.

Unsupervised learning techniques are those that do not require knowledge of the outcome (i.e., they are trained without information as to whether the herd tested positive for BVD) \citep{hahne2008unsupervised}. These include clustering algorithms; by grouping similar data points together, these models can recognise anomalies or outliers that may indicate the outbreak of disease. 

The use of ML in predicting BVD outbreak in Irish cattle herds offers several advantages. Early detection allows for early intervention, reducing the spread of disease. Additionally, ML may help to optimise the use of veterinary resources by identifying which herds are at a higher risk for infection, thus allowing for targeted treatments.
However, the implementation of ML in BVD outbreak prediction is not without its challenges. One significant issue arises due to the balance of response data. The success of the eradication program in Ireland means that, as of 2023, only $0.39\%$ of herds tested positive for BVD infection. The result of this is that $99.61\%$ of the data for 2023 is composed of negative herds, while only $0.39\%$ is composed of positive herds \citep{ahi2024}. This is an issue that must be kept in mind while calibrating and validating the ML models, to ensure accuracy and reliability of results. 

ML offers a promising avenue for enhancing the health management of Irish cattle herds, by enabling the early detection and prediction of BVD. While there are challenges to be addressed, these challenges are outweighed by the potential benefits of such predictive systems, ranging from improved animal welfare to economic gains.

\section{Methods}
\label{methods}

We will divide the ML methods used in this paper into two broad categories, namely binary classification methods and anomaly detection (or one-class classifier) methods. problems involving highly imbalanced data are more suitable for the latter approach, while the former can be tweaked to incorporate such types of data.

\subsection{Binary Classification}
\subsubsection{Generalised Linear Models}
The Generalised Linear Model (GLM) describes a class of models proposed by \citet{nelder1972generalized} that allow for flexible, non-linear relationships  between a response variable $y_{i}$ and predictor variables $x_{i}$.

A GLM has three components. The first is the probability distribution of the response variable. For the purposes of this paper, as we are performing a binary classification task, we implement a logistic regression model, and our response is specified as follows:
$$y_{i} \sim \text{Bernoulli}(\pi_{i}),$$
\noindent where $\pi_i$ is the probability of success (or, in this case, of a herd being BVD positive)

A generalised linear model also has a linear predictor - the linear combination of predictor variables $x$ and coefficients $\beta$ and a link function, which links this linear predictor to the expected value of the Bernoulli distribution, $\pi_i$. The link function associated with logistic regression is the logit function, which maps the values of the linear predictor (which are unbounded) to probability values between 0 and 1:

$$\text{logit}(\pi_{i})=\text{log}\left(\frac{\pi_{i}}{1-\pi_{i}}\right)=\beta_{0}+\beta_{1}x_{1i}+\hdots + \beta_{p}x_{pi},$$

\noindent where $x$ represents known values of predictor variables.

A GLM assumes that the response variables $y_{i}$ are independently distributed. In the case of predicting the presence of BVD in Irish cattle herds, this implies we assume that the disease-status of different herds are independent from one another. In reality, this is likely not the case, since herds that are geographically close to one another may share risk factors, such as shared pastures, wildlife vectors, or farm-to-farm movement of cattle. This spatial dependence suggests that the infection status of one herd may influence the likelihood of infection in neighbouring herds. As a result, failing to account for these dependencies could lead to biased estimates and reduce the accuracy of predictions. In models like GLMs, spatial autocorrelation can be handled by incorporating additional spatial variables or by using more advanced techniques like spatial regression models or Bayesian hierarchical frameworks that explicitly model these dependencies.

\subsubsection{Regularised Regression}
When we perform ordinary least squares (OLS) regression, our aim is to minimise the residual sum of squares (RSS), where the RSS is the sum of the distances between the true response values $y$ and our model predictions $\hat{y}$, 
\begin{equation}
    \text{RSS} = \sum_{i=1}^{n}(y_{i}-\hat{y}_{i})^{2}
\end{equation}

However, OLS makes assumptions about the data that are often not met in practice. In particular, it assumes that multicollinearity between predictors is not present, and that there are more observations than predictors. Real-world datasets often have a large number of predictors, and with a large number of predictors comes the possibility of correlation between them.

The term 'regularised regression' refers to an alternative to OLS modelling, in which the magnitude of the coefficients is reduced, in order to combat these issues and to avoid the model overfitting (an issue that occurs when the model learns the training data to such an extent that it negatively impacts the ability of the model to generalise to new data). The types of regularised regression that we will examine are ridge  \citep{golub1999tikhonov}, LASSO  \citep{tibshirani1996regression} and elastic net regression \citep{zou2005regularization}. 

While all of these models work to avoid overfitting by introducing a penalty term to the loss function, they differ in the choice of this penalty term. Ridge regression introduces a penalty term in the form of the sum of squared coefficients:

\begin{equation}
    \text{RSS}+\lambda\sum_{j=1}^{p}\beta_{j}^2
\end{equation}

\noindent where $\lambda>0$ is the regularisation parameter. If $\lambda=0$, ridge regression reduces to linear regression, and as $\lambda$ increases, the penalty shrinks the coefficients towards zero. A ridge regression model can perform effective regularisation by reducing the magnitude of coefficients for correlated features. However it will not perform feature selection. 

The LASSO (Least Absolute Shrinkage Selector Operator) introduces a penalty term in the form of the sum of the absolute value of the coefficients:

\begin{equation}
    \text{RSS}+\lambda\sum_{j=1}^{p}|\beta_{j}|
\end{equation}

\noindent LASSO regression forces the magnitude of some coefficients to zero, effectively removing certain predictors from the model. As a result, it can perform feature selection, leading to a more parsimonious model. However, when dealing with correlated features, a LASSO model typically selects one feature and discards the others, without preference for which feature is retained  \citep{zou2005regularization}.

Elastic net regression combines the ridge and LASSO penalties: 

\begin{equation}
    \text{RSS}+\lambda_{1}\sum_{j=1}^{p}\beta_{j}^{2}+\lambda_{2}\sum_{j=1}^{p}|\beta_{j}|
\end{equation}
This allows us to benefit from the feature selection provided by the LASSO penalty and the systematic reduction in the magnitude of coefficients for correlated variables provided by the ridge penalty. Choosing the values for the regularisation parameters $\lambda_{1}$ and $\lambda_{2}$ in elastic net regression is critical to balancing the ridge and LASSO penalties effectively. This process typically involves cross-validation to find the values that minimise the model error. Commonly, k-fold cross-validation is used, whereby the dataset is split into k subsets (folds), and the model is trained on $k-1$ folds, and then tested on the remaining fold. This process is repeated for each fold, and the performance is averaged across folds. Various combinations of $\lambda_{1}$ and $\lambda_{2}$ may be tested. The pair of values that maximise model performance - as calculated by a performance metric such as the mean squared error - are chosen as optimal.

\subsubsection{Tree-Based Methods}

\begin{figure}[H]
    \centering
    \includegraphics[width=\textwidth]{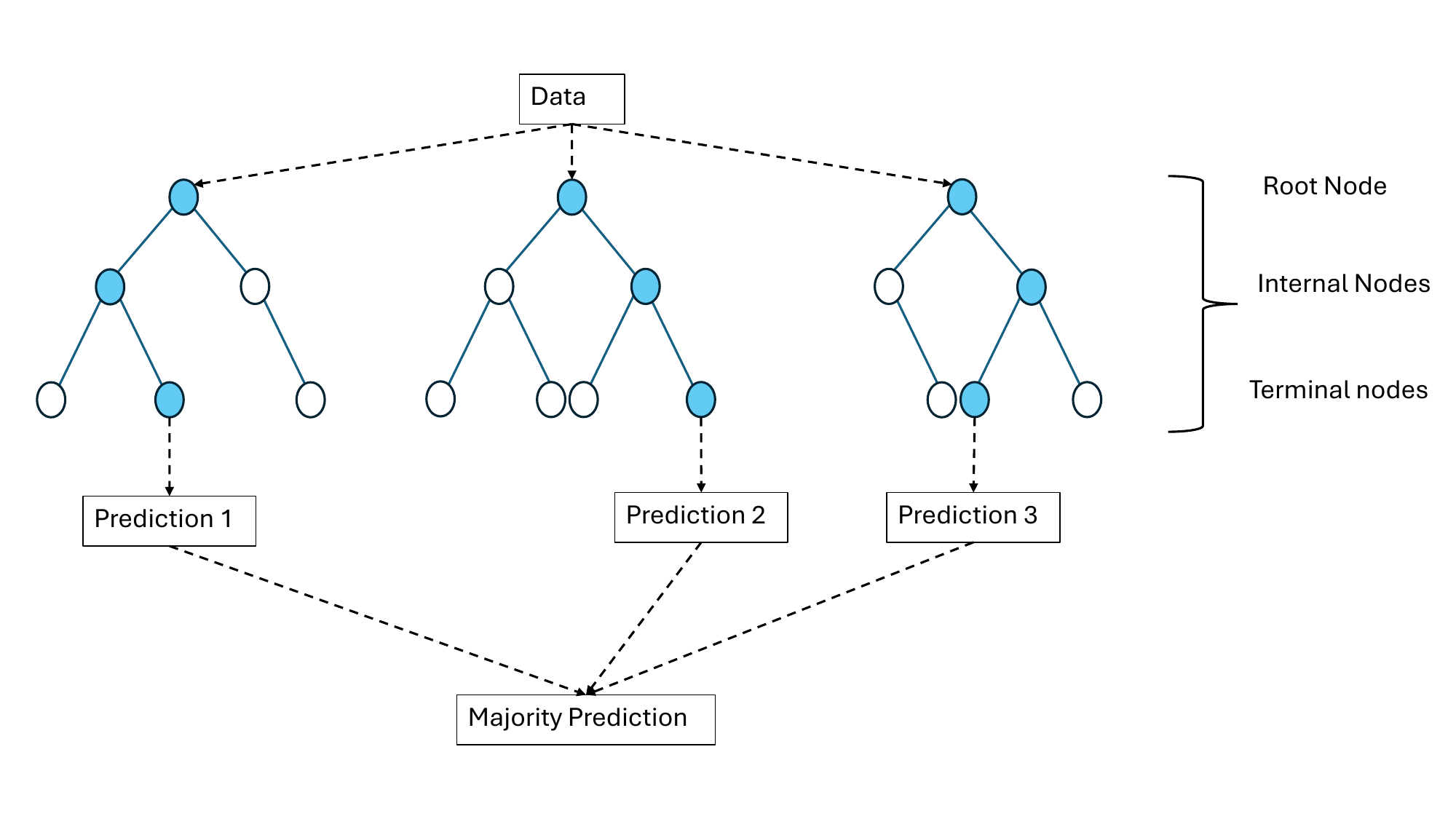}
    \caption{A random forest composed of three trees. Each tree produces a class prediction, and the overall prediction is then determined as the majority of the individual tree predictions.}
    \label{fig:randomforest}
\end{figure}

Classification and Regression Trees (CART)  are a type of decision tree proposed by \citet{breiman1984classification}. They contain root nodes, internal nodes and terminal nodes, all connected by branches, and aim to predict the class of a target variable by learning decision rules. The structure of a random forest composed of three trees is presented in Figure \ref{fig:randomforest}.

A decision tree begins with the root node, calculating for each feature the Gini impurity, which measures the probability of incorrectly classifying an observation if it were labelled based on the class distribution of the dataset. The attribute that provides the smallest Gini impurity is placed at the root. This process is repeated for each node in the tree. Decision trees like CART have the advantage of being easy to interpret. However they are prone to overfitting, and do not generalise well to new data.

Random Forests \citep{breiman2001random} are a machine learning method which combine predictions from an ensemble of decision trees to obtain more stable, accurate predictions than might be obtained using a single decision tree. At each split in a decision tree, a Random Forest examines only a random subset of the available features, and builds smaller trees using those features to avoid overfitting to the training data.

XGBoost (Extreme Gradient Boosting) \citep{chen2016xgboost} is an advanced implementation of the gradient boosting technique, which builds trees sequentially to correct the errors of previous trees. Unlike Random Forests, where trees are grown independently, XGBoost works by adding new trees to the model that focus on areas where the previous trees performed poorly. Each new tree is trained to predict the residuals (errors) of the existing model, improving its predictions in an iterative manner. The optimisation in XGBoost is performed using gradient descent, and the method also introduces techniques such as regularisation (to prevent overfitting), and weighted quantile sketching (to handle large datasets efficiently).

XGBoost has gained popularity due to its high predictive performance and scalability, making it effective for a wide range of machine learning problems. One of its key strengths lies in its ability to handle missing data and noisy datasets while maintaining computational efficiency.

Both Random Forests and XGBoost are ensemble learning techniques, but they differ in how they construct and utilise decision trees. Random Forests rely on bagging (bootstrap aggregating) to reduce variance, while XGBoost uses boosting to reduce both bias and variance, building trees sequentially with gradient descent optimisation.

\subsubsection{Support Vector Machines}
The aim of a Support Vector Machine (SVM) \citep{cortes1995support} is to find a hyperplane that best divides the dataset into two classes. The type of hyperplane used depends on the dimensionality of the data. A dataset with 2 predictors can be separated by a line, a dataset with 3 predictors requires a plane, and a dataset with 4 or more predictors uses a hyperplane.

\begin{figure}[H]
    \centering\includegraphics[width=\textwidth]{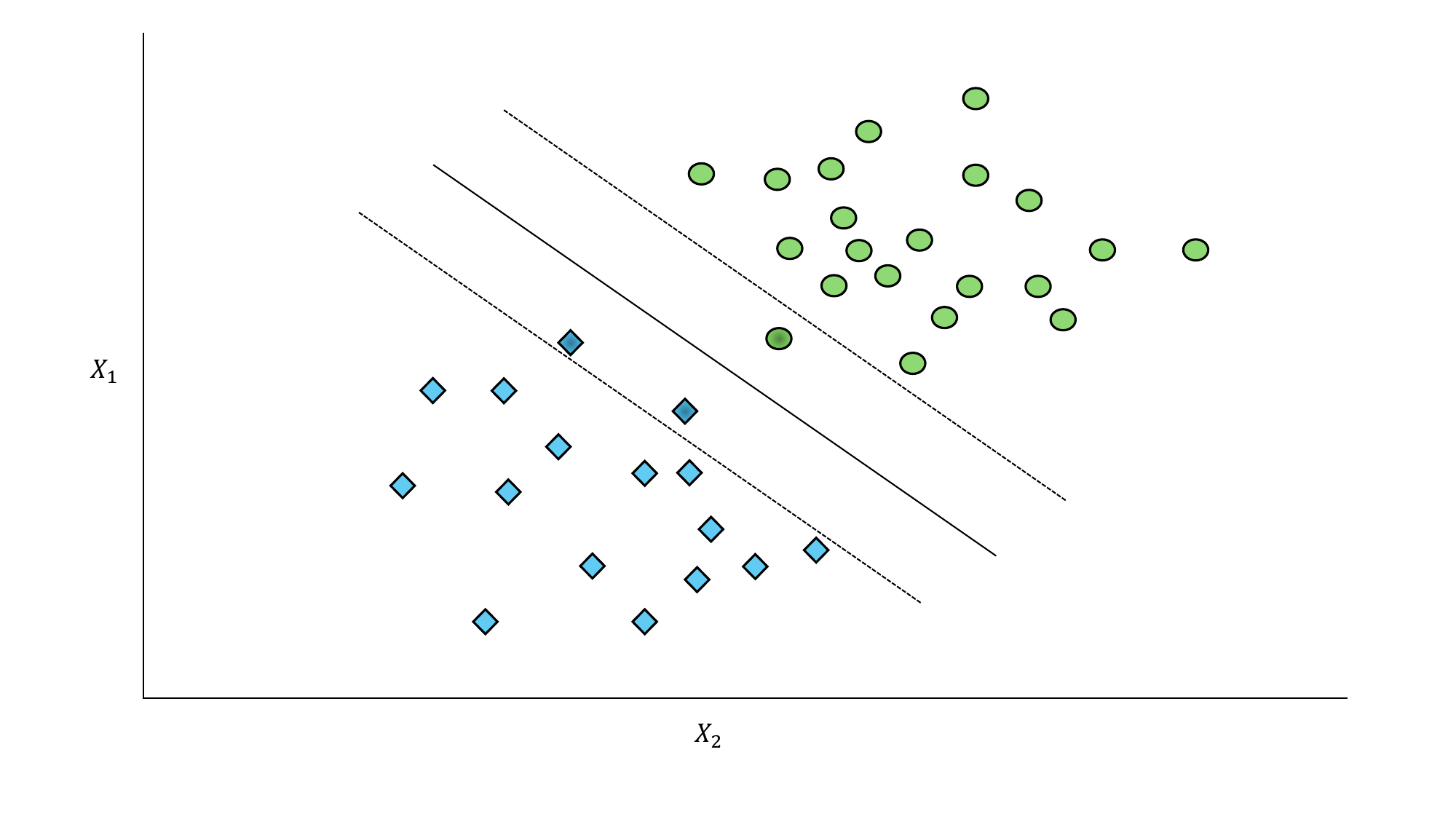}
    \caption{A hyperplane separating two-dimensional data into two classes, where the centre solid line represents the hyperplane, the two dashed lines represent the soft margin, and the shaded observations within the margin are the support vectors.}
    \label{fig:svm}
\end{figure}

A maximal margin classifier attempts to place the hyperplane such that we obtain the largest possible margin. However this type of classifier is very sensitive to outliers in the training data. To address this, we must allow misclassifications to occur. When misclassifications are permitted, the distance between the observations and the hyperplane is called a soft margin. A "support vector" is an observation that is close to the hyperplane and influences how the hyperplane is positioned in space. 
In Figure \ref{fig:svm} we see a hyperplane separating two-dimensional data into two classes. Here the solid line represents the hyperplane, the dashed lines represent the soft margin, and the shaded observations within the margin are the support vectors.

\begin{figure}[H]
    \centering
    \includegraphics[width=\textwidth]{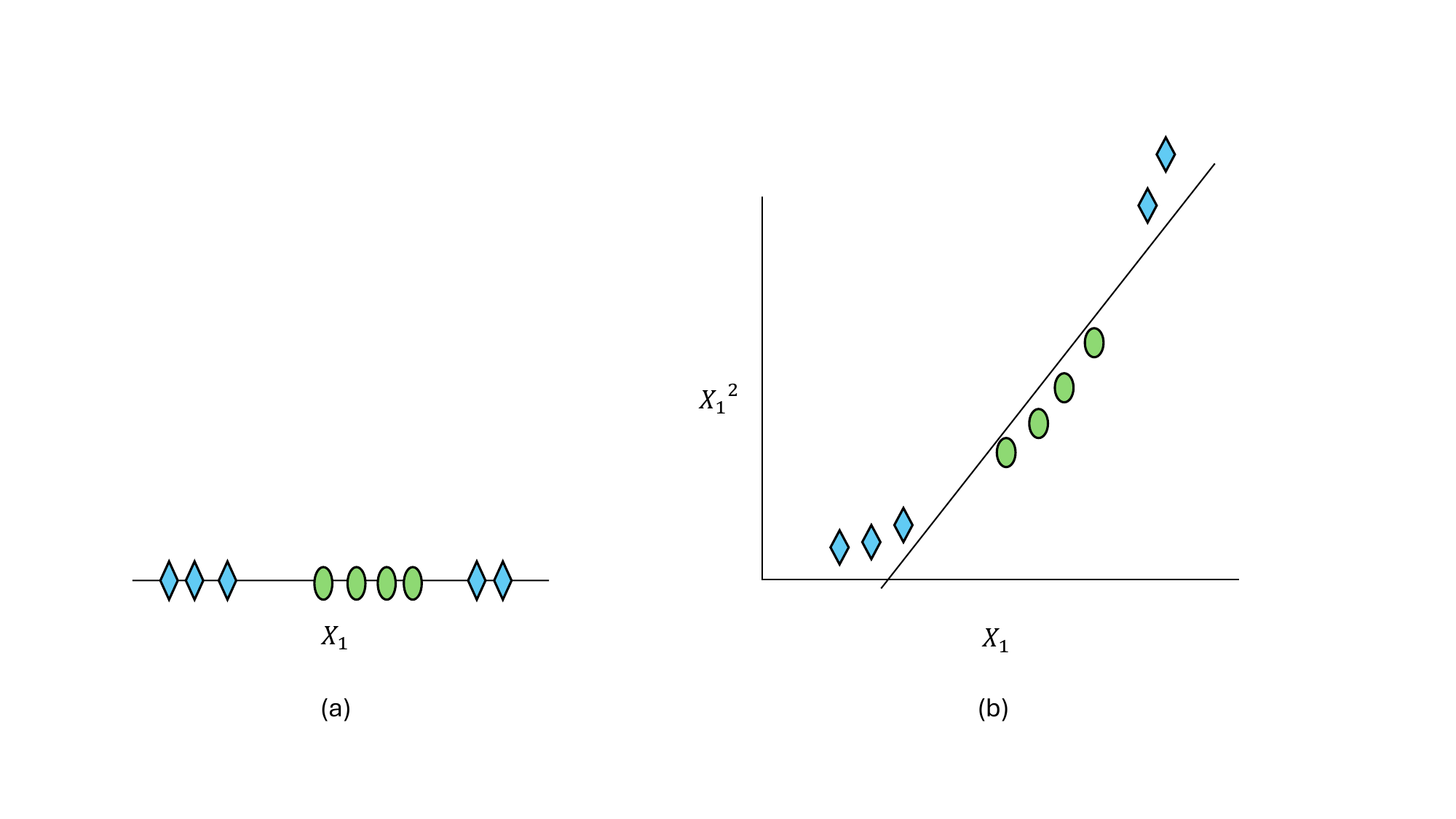}
    \caption{Representation of how a support vector machine may map linear data to a higher dimension to obtain separability. (a) In the original feature space, this two-class data is not linearly separable.  (b) The data is transformed into a higher-dimensional space ($X_{1}^{2}$) to make linear separation of the classes possible. }
    \label{fig:svm2}
\end{figure}

If data is not linearly separable, as in Figure \ref{fig:svm2}(a), we can obtain linear separability by first mapping the data into a higher dimension. In Figure \ref{fig:svm2}(a), the one-dimensional data is not linearly separable. However, if we map it into two dimensions or higher (Figure \ref{fig:svm2}(b)), we may obtain linear separability.

SVMs use kernel functions (functions that transform the non-linear data into higher dimensional linear data) to find support vector classifiers in higher dimensions. There are several possible kernels.  For instance, a polynomial kernel has the form $(a \times b + r)^{d}$, where $a$ and $b$ are observations in the dataset, $r$ is a bias term that shifts the decision boundary in parameter space and $d$ is the degree of the polynomial. Figure \ref{fig:svm2}(b) displays the result of using a polynomial kernel with $d=2$. In contrast, a radial kernel has the form $e^{-\gamma(a-b)^{2}}$, where $\gamma$ scales the influence that observations $a$ and $b$ can have on one another. This kernel finds support vector classifiers in infinite dimensions. It behaves like a weighted nearest neighbours algorithm, whereby closer observations have a high degree of influence classifying new observations, while observations further away are less influential. 

\subsection{Anomaly Detection}
\subsubsection{Local Outlier Factor}
Local Outlier Factor (LOF) is an anomaly-detection algorithm proposed by \citet{breunig2000lof}. LOF is based on a concept of a local density, where density is estimated by the distance between the observation in question and k-nearest neighbours. Estimating the local density of each observation allows us to make comparisons, and observations with a substantially lower local density than their neighbours are considered to be outliers.  

\begin{figure}[H]
    \centering
    \includegraphics[width=0.5\linewidth]{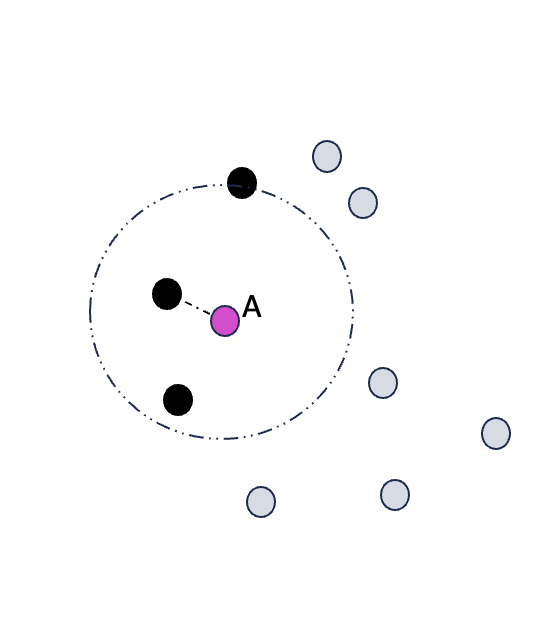}
    \caption{Representation of the k-neighbourhood of observation A (highlighted in pink), where the shaded observations represent the three nearest neighbours, and the dashed circle represents the k-neighbourhood $N_{3}(A)$.}
    \label{fig:LOF_Kdist}
\end{figure}

Let k-distance(A) be the distance from observation A to the $k^{th}$ nearest neighbour, an $N_{k}(A)$ be the k-neighbourhood or the set of k-nearest neighbours. Figure \ref{fig:LOF_Kdist} shows the point A with its three nearest neighbours highlighted, where the dashed circle represents $N_{3}(A)$. 

\begin{figure}[H]
    \centering
    \includegraphics[width=\linewidth]{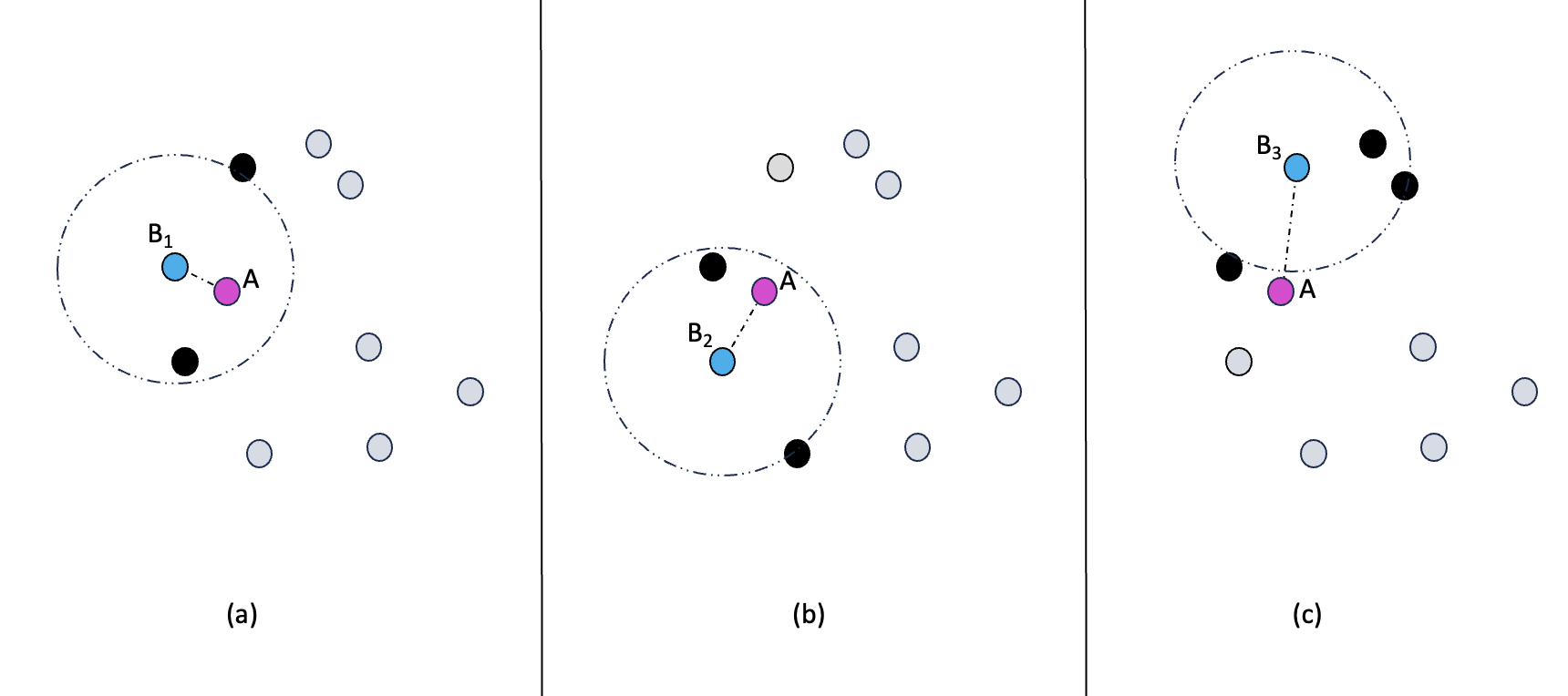}
    \caption{The RD of observation A (highlighted pink) to each of its three nearest neighbours (highlighted blue). The nearest neighbours to observations $B_{1}$,$B_{2}$,and $B_{3}$ are shaded, while the k-neighbourhood for observations $B_{1}$,$B_{2}$,and $B_{3}$ is represented with by a dashed circle.}
    \label{fig:RD}
\end{figure}

The reachability distance (RD) uses the k-distance and is defined as follows:

$$RD(A,B)=\text{max}(\text{k-distance}(B), d(A,B))$$

\noindent i.e., the RD of observation A from observation B is the distance $d(A,B)$ from observation A to observation B, but must be at least the k-distance(B), because observations within the k-neighbourhood of B are considered to have an equal distance from B. For example,  in Figure \ref{fig:RD}, we see the RD of observation A to each of its three nearest neighbours, $B_{1}$, $B_{2}$ and $B_{3}$. Here the shaded observations and dashed circle are the nearest neighbours and neighbourhood $N_{3}$ respectively for observation B, rather than observation A. In Figure \ref{fig:RD}(a) and Figure \ref{fig:RD}(b), there are two shaded observations, along with observation A. This is because, for $B_{1}$ and $B_{2}$, observation A is within their $N_{3}$. However, for Figure \ref{fig:RD}(c), there are three shaded points. This is because observation A is not one of the three nearest neighbours to $B_{3}$, and so is not contained within its $N_{3}$. 
The result is that $RD(A, B_{1})=\text{k-distance}(B_{1})$ and $RD(A, B_{2})=\text{k-distance}(B_{2})$, while $RD(A, B_{3})=d(A,B_{3})$.

The local reachability density (LRD) of A is defined as the inverse of the mean RD of the observation A from its neighbours. 

$$LRD(A)=\Bigg(\frac{\sum_{B \in N_{k}(A)}RD(A,B)}{|N_{k}(A)|}\Bigg)^{-1}$$

The local outlier factor is then computed as the average of the ratio of the local reachability density of A and the reachability densities of A's nearest neighbours. An outlier may be identified by a local outlier factor value greater than $1$. 

$$LOF(A)=\frac{\sum_{B \in N_{k}(A)}\frac{LRD(B)}{LRD(A)}}{|N_{k}(A)|}$$

\noindent if $LOF(A) > 1$, the local density of point A is lower than its neighbours, indicating that it is an outlier. Conversely, if $LOF(A) \approx 1$, the density is comparable to its neighbours, meaning the point behaves similarly to them and is likely not an outlier.

\subsubsection{Angle-Based Outlier Detection}
\begin{figure}[H]
    \centering
    \includegraphics[width=\textwidth]{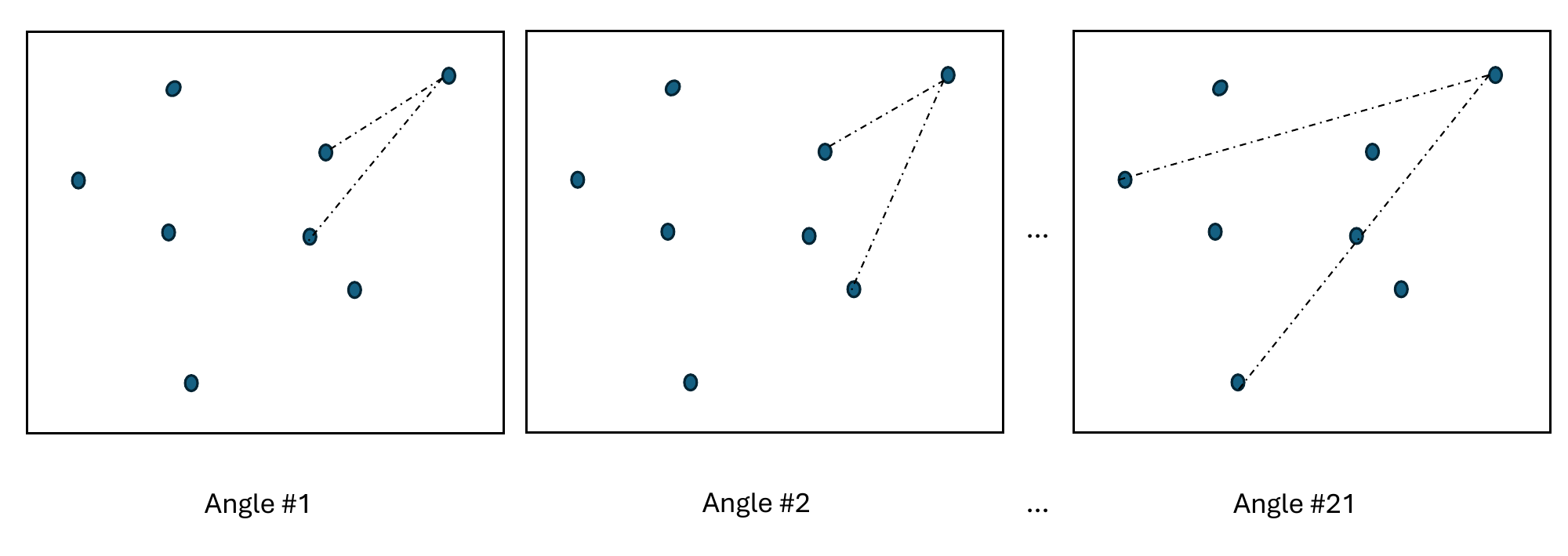}
    \caption{The angles are calculated between each observation and every other pair of observations (21 total angles per observation for this dataset), and the variance of these angles are then calculated to determine the angle-based outlier factor.}
    \label{fig:abod}
\end{figure}

Angle-Based Outlier Detection is an anomaly-detection algorithm proposed by \citet{kriegel2008angle}. For each observation, the angle-based outlier factor (ABOF) is determined by first calculating the angle between each observation and every other pair of observations in the dataset, and then calculating the variance of these angles, weighted by the distance between points. Figure \ref{fig:abod} presents the calculation of these angles, using data in two dimensions.  

Given a collection of observations $D$, where three observations $A$, $B$, $C \in D$, the angle-based outlier factor for observation $A$ is as follows:

$$\text{ABOF}(A)=\text{Var}\Bigg(\frac{\langle AB ,  AC \rangle}{||AB||^{2}\cdot||AC||^{2}}\Bigg)$$
where $AB$ denotes the vector created by connecting  observation $A$ to observation $B$, the scalar product is denoted by $\langle . , . \rangle$, and $||AB||$ represents the magnitude of the vector $AB$. Outliers are identified as observations with ABOF values below a certain threshold, where the threshold value is determined using cross-validation.

The ABOF method is not restricted to two-dimensional data. For datasets with $p>2$ predictors, the angles are generalised to higher-dimensional vectors, where the angle between vectors is computed using the dot product in p-dimensional space. This generalisation allows ABOF to be applied to multivariate datasets. For example, if we were working with data in 3-dimensional space, and had observations $A$, $B$ and $C$, each with three coordinates, corresponding to three predictors: $A=(A_{1}, A_{2},A_{3})$, $B=(B_{1}, B_{2},B_{3})$ and $C=(C_{1}, C_{2},C_{3})$. Vectors may be created in space between these observations (e.g., $AB = (B_{1}-A_{1}, B_{2}-A_{2}, B_{3}-A_{3})$. The angle-based outlier factor is then calculated as before. 

\subsubsection{Mahalanobis Distance Classification}
The Mahalanobis distance \citep{mahalanobis1936generalised} is a distance metric that computes the distance between a point and a distribution, by considering how many standard deviations the point is from the distribution. The Mahalanobis distance formula is given as:
$$D^{2}=(x-\mu)^{T}\cdot \Sigma^{-1}\cdot(x-\mu)$$
where $x$ is the observation whose distance we want to compute, $\mu$ is the mean vector and $\Sigma$ is the covariance matrix.

Outlier detection using the Mahalanobis distance involves calculating the Mahalanobis distance between each observation of interest and the centre of the distribution. The Mahalanobis distance follows a chi-square distribution when the data is multivariate normally distributed. A cutoff point is thus determined using the chi-square distribution with degrees of freedom equal to the number of variables in the dataset. To establish the cutoff point, we can specify a significance level (e.g., $\alpha=0.05$), which indicates the probability of incorrectly identifying an observation as an outlier.  Observations with a Mahalanobis distance is greater than this cutoff are identified as outliers. For example, if we had a dataset with three variables (thus three degrees of freedom) and choose a significance level of 0.05, we would look up the chi-square value for 0.95 probability (1 - 0.05) in the chi-square table with three degrees of freedom. Observations with a Mahalanobis distance exceeding this value would be flagged as outliers. Alternative approaches to choosing a cutoff point include using the training data to determine an optimal threshold. This may be done by analysing the distribution of Mahalanobis distances of both normal observations and outliers, and identifying an appropriate cutoff that balances false positives and false negatives.

\subsubsection{Minimum Covariance Determinant}
The Minimum Covariance Determinant (MCD) \citep{rousseeuw1999fast} operates in a similar manner to the Mahalanobis distance classifier. The difference lies in the data used to determine the mean vector $\mu$ and covariance matrix $\Sigma$. The Mahalanobis distance classifier estimates $\mu$ and $\Sigma$ of the distribution using the entire dataset. However, if there are outliers in the dataset, this may skew these parameter estimates, and this will in turn make outliers appear less anomalous than they are, possibly leading to outliers being incorrectly identified as non-anomalous observations. 

MCD subsamples the dataset with the objective of finding a subsample that does not contain any outliers. It does this by randomly subsampling the data, and computing $\mu$ and $\Sigma$ for each subsample. MCD selects the subsample least likely to contain outliers by determining how densely distributed each subsample is. The presence of outliers in a subsample will reduce the density of that subsample. The determinant of the covariance matrix measures the width of the distribution, and so the determinant of $\Sigma$ for each subsample is calculated, and the estimates are retained for the subsample whose covariance matrix had the smallest determinant.

\subsubsection{Isolation Forests}
Similar to a random forest, and isolation forest is an ensemble method, i.e., it uses the average of the predictions by several decision trees when assigning an anomaly score to an observation. 

\begin{figure}[H]
    \centering
    \includegraphics[width=\textwidth]{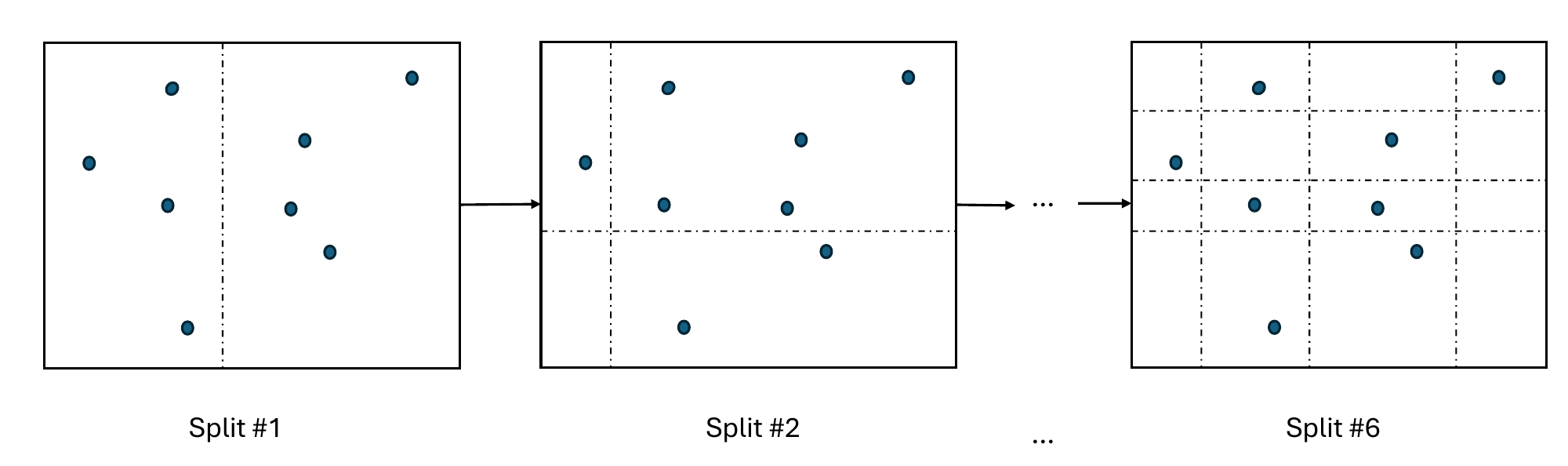}
    \caption{An isolation forest for two-dimensional data. The isolation forest randomly splits the data along a random dimension, and continues to do so until every observation is isolated in a terminal node. In this case, six splits were required to isolate all observations.}
    \label{fig:isolation_forest}
\end{figure}

An isolation forest selects a random feature, and randomly splits the dataset along that feature. This creates two subspaces, one on either side of the split. All observations fall into one of the two subspaces. This process (randomly selecting a feature and randomly splitting along that feature) is repeated until no further split is possible (every observation is contained within its own terminal node). This process of splitting data is presented in Figure \ref{fig:isolation_forest}.

Isolation forests identify outliers based on the number of splits required to isolate an observation in its own terminal node. It will normally require fewer splits to isolate an outlier than to isolate a normal observation. In Figure \ref{fig:isolation_forest}, the observation isolated at the second split is more likely to be an outlier observations isolated later in the process. 
For each observation, the anomaly score is calculated as the mean number of splits required for isolation. To detect outliers, a threshold is set for the anomaly score. Observations with scores below this threshold are considered normal, while those above it are classified as outliers. The choice of threshold can be based on domain knowledge, or through cross-validation to optimise detection performance.

\subsubsection{Autoencoders}
An autoencoder is a type of unsupervised neural network that aims to take the input values, extract the essential information, and use this information to re-create these input values. It does this by combining an encoder with a decoder. An autoencoder, like all neural networks, has an input layer, a number of hidden layers, and an output layer. In an autoencoder, the hidden layers must have fewer dimensions than those of the input or output layers.

\begin{figure}
    \centering
    \includegraphics[width=\textwidth]{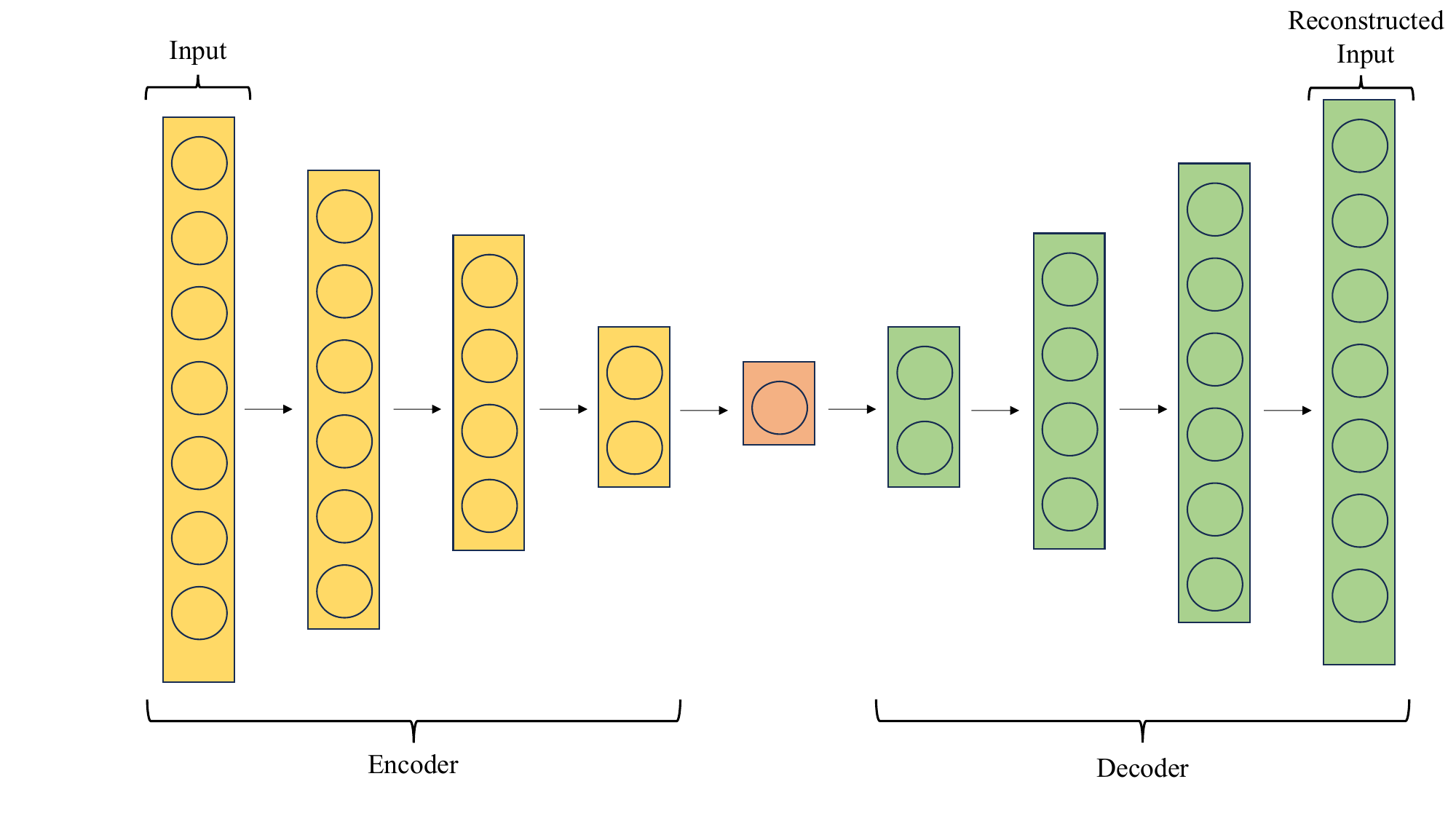}
    \caption{The structure of an autoencoder consists of an encoder which compresses the input data, an a decoder which then reconstructs the input data.}
    \label{fig:autoencoder}
\end{figure}
Figure \ref{fig:autoencoder} shows the encoder and decoder. Input values are accepted in the input layer, and through the use of subsequent hidden layers, each smaller than the last, the encoder compresses the input values via linear and non-linear operations. The decoder then mirrors this process, reconstructing the information until we reach the output layer, with a dimension equal to that of the input layer. 

The output of this autoencoder is then a reconstruction of the input values. If an autoencoder is trained using just the ``normal" observations (in our case, the BVD negative herds), and then tested using data that contains outliers, the autoencoder should struggle to reproduce these observations. For each observation, we can assign a reconstruction error which measures the difference between the reconstructed and observed values. A threshold may be chosen via cross-validation, and any observation with an associated reconstruction error above this threshold may be identified as an outlier. 

\section{Techniques for Imbalanced Data}
\label{imbalanced_data}
\subsection{Evaluation Metrics}
Because of the imbalance present in the data, the evaluation metrics must be carefully chosen. Because there are far greater numbers of negative BVD cases than positive ones, evaluation metrics must be chosen that will highlight when these positive BVD cases are not being correctly predicted. 

The positive predictive value (PPV) represents the total number of true positive predictions (herds which were BVD positive that were correctly predicted to be BVD positive) as a fraction of the total number of positive predictions (including the true positive predictions described above and the false positive predictions -- those herds that were negative for BVD but were predicted to be positive for BVD). Maximising the PPV means maximising the number of true positive predictions while minimising the number of false positive predictions. The PPV can be written as 

\begin{equation*}
    \text{Positive Predictive Value} = \frac{\text{True Positives}}{\text{True Positives} + \text{False Positives}}
\end{equation*}

Similarly, diagnostic test sensitivity represents the number of true positive predictions as a fraction of the total number of positive herds, both those predicted to be BVD-positive and those predicted to be BVD-negative. Maximising the sensitivity means maximising the number of true positive predictions while minimising the number of false negative predictions. The sensitivity can be written as:

\begin{equation*}
    \text{Sensitivity} = \frac{\text{True Positives}}{\text{True Positives} + \text{False Negatives}}
\end{equation*}

Overall model predictive performance can be measured using the $\text{F}_{1}$-score. This is the harmonic mean of the PPV and the diagnostic test sensitivity, and so maximising the $\text{F}_{1}$-score means maximising both the PPV and diagnostic test sensitivity simultaneously. The $\text{F}_1$ score is given by: 

\begin{equation*}
    F_1=\frac{2 \cdot \text{Positive Predictive Value} \cdot \text{Sensitivity}}{\text{Positive Predictive Value} + \text{Sensitivity}}
\end{equation*}

Another important evaluation metric is the Area Under the Receiver Operating Characteristic Curve (AUC-ROC) \citep{hanley1982meaning}. This metric is particularly useful in imbalanced datasets, where the number of negative cases far exceeds the number of positive cases.

The ROC curve is a plot of the Sensitivity against the False Positive Rate (FPR), which is defined as the fraction of negative cases that are incorrectly predicted to be positive:

\begin{equation*} \text{False Positive Rate} = \frac{\text{False Positives}}{\text{False Positives} + \text{True Negatives}} \end{equation*}

The AUC represents the probability that the model will rank a randomly chosen positive instance higher than a randomly chosen negative one. A model with an AUC of 0.5 performs on par with a random choice, while a model with an AUC of 1.0 makes perfect predictions. The AUC is especially valuable in imbalanced datasets, as it evaluates the model’s ability to discriminate between positive and negative cases, regardless of the threshold chosen for classification.

Because the AUC is independent of the chosen classification threshold, it can give a more holistic view of model performance than metrics like PPV and Sensitivity, which depend on a fixed threshold. This makes it a robust tool for evaluating models where the cost of false positives and false negatives must be balanced carefully.

\subsection{Weighted Classification}
When performing multi-class classification, beyond choosing the correct evaluation metric, there are a few approaches we can take to improve the predictive ability of the model. The first is through the use of class weights and performance of weighted classification. A weight is assigned to every observation in the dataset, and higher weights are assigned to observations of the minority class, which increases the cost of incorrectly classifying that class.  

A commonly used approach to assigning weights -- and the approach that is implemented here -- is assigning weights that are inversely proportional to the frequency of the class, i.e.,
\begin{equation*}
    \text{weight(class A)}=\frac{\text{Total number of observations}}{\text{(Number of classes)}\text{(Number of observations in class A)}}
\end{equation*}

For example, if the data contains 100 observations, 90 of which belong to the BVD-negative class and 10 of which belong to the BVD-positive class, then a weight of 0.556 would be applied to the BVD-negative class, while a much greater weight of 5.0 would be applied to the BVD-positive class.

This results in a model that is more sensitive to the minority class. During model training, the class weights are multiplied by the loss function. The result is that errors made in classifying the minority class are more heavily penalised than errors made in classifying the majority class. This aids the model in representing the imbalanced nature of the dataset. 

\subsection{Resampling Techniques}
An alternative approach to dealing with imbalanced data is to under-sample (randomly remove observations from the majority class) or over-sample (randomly duplicate observations from the minority class) the training data. This results in a dataset that is less imbalanced than the original dataset. 

\begin{figure}[H]
    \centering
    \includegraphics[width=\textwidth]{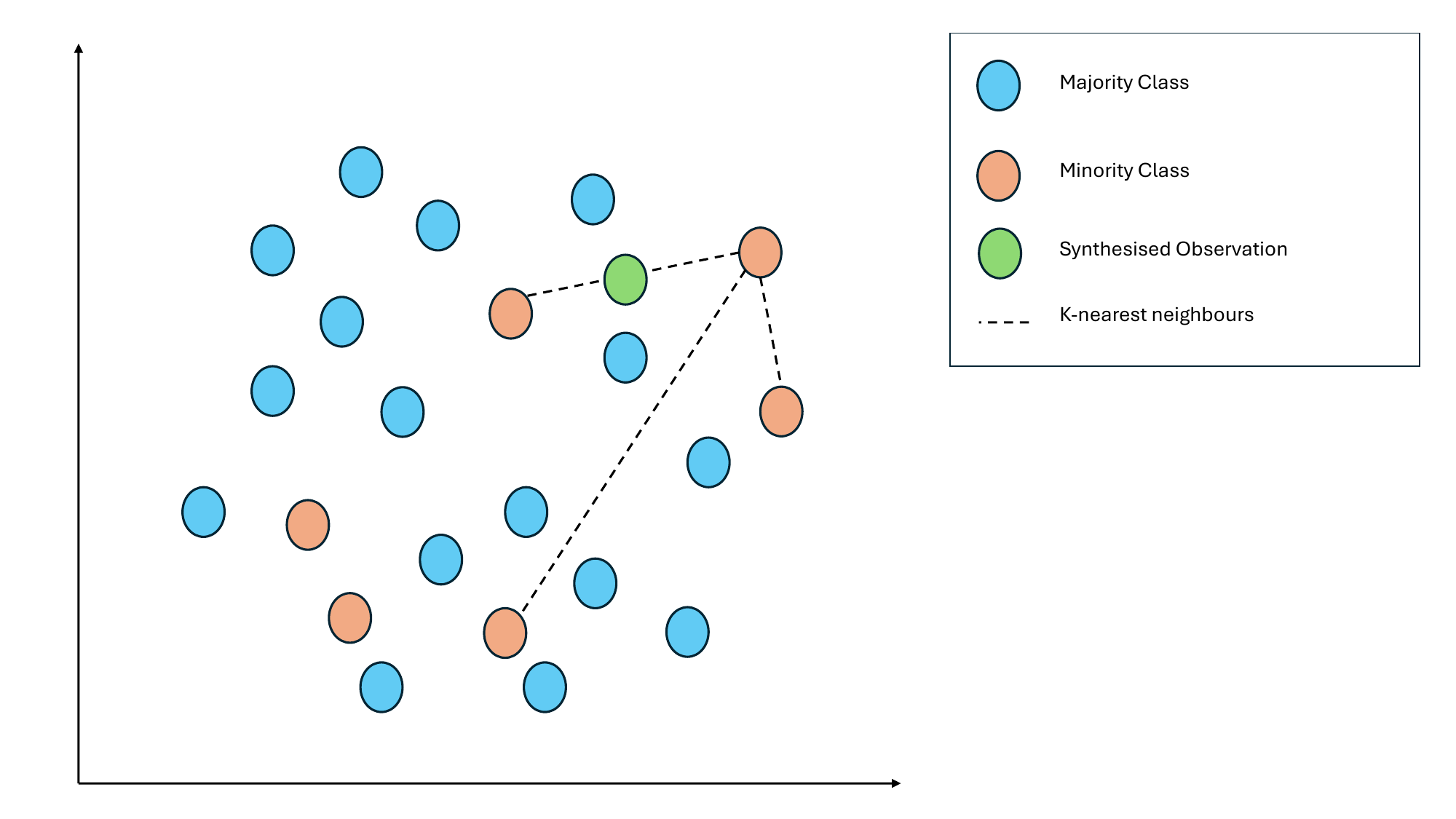}
    \caption{Synthetic Minority Oversampling Technique randomly selects a point from the minority class and computes its k-nearest neighbours. It then selects one of those neighbours and creates a new observation at a point between the two observations.}
    \label{fig:smote_diagram}
\end{figure}

In this study we examine a form of over-sampling called SMOTE (Synthetic Minority Oversampling TEchnique) \citep{chawla2002smote}. This technique involves creating synthetic observations from the minority class, by randomly picking an observation from the minority class and finding its k-nearest (minority class) neighbours. One of these neighbours is then chosen at random and a new observation is synthesised at a randomly selected point between the two examples in feature space. The recommended approach to over-sampling using SMOTE is to first randomly remove some of the majority class observations via random under-sampling, and to subsequently use SMOTE to over-sample the minority class, so that class distributions are balanced. This technique is illustrated in Figure \ref{fig:smote_diagram}.

\section{Simulation Study}
\label{simulation}
In this section we describe simulation studies which were conducted with the aim of determining the relative ability of various machine learning techniques to predict BVD re-emergence. To this end, data was simulated and machine learning methods were fitted in R \citep{R}. The data simulation was carried out with the objective of producing data with a structure similar to the true data. For $i \in \{1,\ldots,R\}$ herds and $j \in \{1,\ldots,J\}$ years, whether herd $i$ tests positive for BVD in year $j$ is determined as follows:

\begin{equation*}
    Y_{i,j} = \begin{cases}
  \text{Positive}  & \text{if } Z_{i,j} \geq Q \\
  \text{Negative} & \text{otherwise}
\end{cases}
\end{equation*}

Here, $Q$ is a quantile that determines the proportion of positive herds in the population. We wish to assess the affect of varying class balances on model predictive performances, so we vary $Q \in (0.5, 0.1, 0.05, 0.01)$.
To simulate our binary response data $Y_{i,j}$ we use a latent variable $Z_{i,j}$ to capture the combined influence of various factors on the disease status of a herd. $Z_{i,j}$ is a continuous measure that reflects the underlying risk or propensity for each herd to develop BVD, based on the effect of environmental and/or population-level parameters, assumed as

\begin{equation*}
    Z_{i,j} \sim \text{Normal}(\mu_{i,j})
\end{equation*}

\noindent where the mean of this latent variable $\mu_{i,j}$ is determined by:

\begin{equation*}
\begin{aligned}
     \mu_{i,j} = &\beta_{0}+\beta_{1}x_{1}+\beta_{2}x_{2}+\beta_{3}x_{3}+\beta_{4}x_{4}+\beta_{5}x_{5}+\beta_{6}x_{6}+\beta_{7}x_{7}+\\
     &\beta_{8}\text{sin}(\pi (x_{5_{j-1}})(x_{6_{j-1}}))+\beta_{9}(x_{8}-(\text{min}(x_{8})+\text{max}(x_{8}))/{2})^{2}\\
\end{aligned}
\end{equation*}

\noindent where $x_{1}=Y_{i,j-1}$ represents whether a herd tested positive for BVD the previous year, while $x_{2}=Y_{i,j-2}$ and $x_{3}=Y_{i,j-3}$ represent whether a herd tested positive for BVD two years and three years previously, respectively. $x_{4}$ represents the number of animals a herd imported in the previous year, $x_{5}$ represents the number of stillborn animals in a herd in the previous year, $x_{6}$ represents the number of animals moved to the knackery from a herd in the previous year, $x_{7}$ represents the size of the herd in the previous year, and $x_{8}$ represents the local disease density, calculated as the number of neighbouring herds that tested positive for BVD in the previous year, divided by the total number of neighbouring herds. The variables $x_{1}$ to $x_{7}$ have a straightforward, linear relationship with the outcome. This reflects the idea that, for instance, a greater number of stillborn animals or higher disease density likely have a direct, proportional impact on herd health. 

In addition to the linear effects of $x_{1}$ to $x_{7}$, more complex relationships are also incorporated. The term $\text{sin}(\pi (x_{5_{j-1}})(x_{6_{j-1}}))$ represents an interaction between the number of stillborn individuals and the number of individuals moved to the knackery. The use of the sine function introduces a periodic, wave-like relationship between these two variables, resulting in a combined effect that is not simply additive or linear. This might model a scenario where specific combinations of stillbirths and animals moved to the knackery lead to disproportionately large or small effects on herd health, possibly because these factors together could indicate broader health or management issues within the herd and could reflect a complex biological or environmental interaction that isn't captured by simple linear terms. Additionally, the term $(x_{8}-(\text{min}(x_{8})+\text{max}(x_{8}))/{2})^{2}$ accounts for how deviations in the herd size from a certain average may have a squared effect, which might reflect increased risks or stresses associated with unusually small or large herds. 

By combining both linear and more complex effects, this simulation study aims to determine how well machine learning methodologies are able to capture a range of relationships between possible factors and the BVD-status of a herd, providing a more realistic model of how various influences might interact in the real world.

 To assess the effect of sample size on disease prediction, two scenarios were examined. The first was a scenario meant to represent a small sample size, with data on $800$ herds. The second is a scenario with a larger sample size of $10,000$ herds. These sample sizes were chosen to mimic real data, where one county might be expected to contain several hundred herds, and multiple counties examined together would be expected to contain several thousand herds. Additionally, where appropriate, the effect on BVD prediction of oversampling via SMOTE was examined. Where the ML techniques allowed, 5-Fold Cross Validation (CV) was carried out. The predictive accuracy of each model was assessed using the positive predictive value, the diagnostic test sensitivity, the F1-score, and the area under the ROC curve (AUC) value.

In Table \ref{tab:model_table1} we present the results of the implementation of 13 machine learning techniques on data that has been resampled using SMOTE. Table \ref{tab:model_table1}(a) contains model results for data with balanced classes ($50\%$ of observations represent herds that are positive for BVD and $50\%$ represent herds that are negative for BVD). While Table \ref{tab:model_table1} contains results for data that has been resampled using SMOTE, this resampling would not have had an effect on the results for Table \ref{tab:model_table1}(a), as these datasets already contained balanced classes. This sub-table is further divided into results for a small sample size (800 herds) and results for a large sample size (10,000) herds. At both small and large sample sizes, the Random Forest produces the most accurate predictions, with high sensitivity and positive predictive value. This means that the Random Forest is predicting "true positives" (herds that are truly positive for BVD, and are predicted correctly) with a high degree of accuracy, while avoiding "false positives" (herds that are truly negative for BVD but are predicted positive) and "false negatives" (herds that are truly positive for BVD but are predicted negative). The Support Vector Machine also produces accurate predictions, with high sensitivity and positive predictive values. While the XGBoost model has very high positive predictive value, it has low sensitivity, which means that there is an issue with false negative predictions for this technique. The remaining techniques (Generalised Linear Models, LASSO regression, Ridge regression and Elastic Net regression) fail to correctly predict BVD cases, with AUC values of approximately 0.5, which is equivalent to the model producing BVD predictions through random guessing. 

\begin{table}[H]
    \centering
    \begin{tabular}{ c| c c c c | c c c c }
    \hline 
    \multicolumn{9}{c}{(a) $50\%$ Positive Herds}\\
    \hline
     & \multicolumn{4}{|c}{Small Sample Size}& \multicolumn{4}{|c}{Large Sample Size}\\
    Model  & PPV & Sensitivity & F1 & AUC   & PPV & Sensitivity & F1 & AUC\\ 
    \hline
    GLM & 0.486 & 0.483 &0.484&0.532&0.504 &0.494 &0.499 &0.499\\
    Random Forest & 0.995 & \textbf{0.993}&0.994&\textbf{0.999}&0.998 &\textbf{1.000} &0.999 &\textbf{1.000}\\
    LASSO & 0.500 & 0.475&0.485&0.537& 0.512&0.588 &0.547 &0.517\\
    Ridge & 0.476 & 0.460&0.468&0.548& 0.505&0.528 &0.516 &0.509\\
    Elastic Net& 0.481 & 0.467&0.474&0.544& 0.502&0.591 &0.543 &0.502\\
    SVM &0.891&0.887&0.889&0.961&0.951&0.940&0.945&0.986\\
    XGBoost &1.000&0.425&0.596&0.990&0.996&0.509&0.674&0.998\\
 
    \hline
     \multicolumn{9}{c}{(b) $10\%$ Positive Herds}\\
     \hline
     & \multicolumn{4}{|c}{Small Sample Size}& \multicolumn{4}{|c}{Large Sample Size}\\
    Model  & PPV & Sensitivity & F1 & AUC   & PPV & Sensitivity & F1 & AUC\\ 
    \hline
    GLM & 0.156 & 0.287&0.202&0.556&0.105 &0.465 &0.172 &0.507\\
    Random Forest & 0.987 & 0.980&0.985&\textbf{0.997}&1.000 &0.998 &0.999 &\textbf{1.000}\\
    LASSO &0.117  &0.350 &0.175&0.505& 0.107&0.503 &0.177 &0.524\\
    Ridge & 0.101 &0.401 &0.161&0.509& 0.109&0.513 &0.181 &0.531\\
    Elastic Net& 0.101 & 0.403&0.162&0.511& 0.116&0.539 &0.191 &0.538\\
    SVM &0.459&0.712&0.558&0.915&0.571&0.928&0.706&0.981\\
    XGBoost &0.970&\textbf{1.000}&0.985&0.992&0.979&\textbf{1.000}&0.994&0.999\\
    \hline

    \multicolumn{9}{c}{(c) $5\%$ Positive Herds}\\
    \hline
     & \multicolumn{4}{|c}{Small Sample Size}& \multicolumn{4}{|c}{Large Sample Size}\\
    Model  & PPV & Sensitivity & F1 & AUC   & PPV & Sensitivity & F1 & AUC\\ 
    \hline
    GLM & 0.047  & 0.125&0.068&0.484&0.054 &0.044 &0.096 &0.532\\
    Random Forest & 0.975  & \textbf{1.000}&0.987&\textbf{1.000}&0.986 &\textbf{1.000} &0.993 &\textbf{0.999}\\
    LASSO & 0.054 & 0.999&0.095&0.502& 0.051&0.999 &0.090 &0.500\\
    Ridge & 0.051 & 0.481&0.091&0.523& 0.052&0.476 &0.094 &0.522\\
    Elastic Net& 0.050 & 0.999&0.092&0.500& 0.050&0.999 &0.095 &0.500\\
    SVM &0.348&0.750&0.476&0.841&0.399&0.884&0.551&0.969\\
    XGBoost &0.900&0.920&0.900&0.991&0.974&0.980&0.987&0.994\\
    \hline

    \multicolumn{9}{c}{(d) $1\%$ Positive Herds}\\
    \hline
     & \multicolumn{4}{|c}{Small Sample Size}& \multicolumn{4}{|c}{Large Sample Size}\\
    Model  & PPV & Sensitivity & F1 & AUC   & PPV & Sensitivity & F1 & AUC\\ 
    \hline
    GLM &0.000  &0.000 &-&0.493&0.012 &0.024 &0.370 &0.539\\
    Random Forest&0.125  & 0.125&0.125&\textbf{0.796}& 1.000 &0.980 &0.989 &\textbf{0.998}\\
    LASSO & 0.029 & 0.375&0.054&0.539& 0.011&0.999 &0.019 &0.500\\
    Ridge & 0.016 & 0.250&0.031&0.565& 0.014&0.500 &0.027 &0.553\\
    Elastic Net& 0.021 & 0.375&0.039&0.564& 0.010&0.999 &0.019 &0.500\\
    SVM &0.016&0.250&0.031&0.487&0.137&0.700&0.230&0.780\\
    XGBoost &0.600&\textbf{0.500}&0.545&0.746&0.687&\textbf{1.000}&0.814&0.997\\
    \hline
    \end{tabular}
    \caption{Classifier results for (a) $50\%$ positive herds, (b) $10\%$ positive herds, (c) $5\%$ positive herds, and (d) $1\%$ positive herds, using data that has been resampled with SMOTE. In each case, the model with the highest AUC, and the model with the highest sensitivity are highlighted.}
    \label{tab:model_table1}
\end{table}

Table \ref{tab:model_table1}(b) contains modelling results for data with a class imbalance of $90\%$ BVD-negative herds and $10\%$ BVD-positive herds, which was then resampled using SMOTE, to obtain a balance between the classes. At both small and large sample sizes, the most accurate herd predictions can again be obtained using a Random Forest, with an AUC value of $0.997$ and $1.00$ respectively. The XGBoost model in this scenario also produces accurate BVD predictions, as indicated by a sensitivity value of $1.00$ for both small- and large-scale simulations, which indicates that this model did not produce a single false negative prediction (i.e., all herds that tested positive for BVD were correctly predicted as positive for BVD). The Support Vector Machine, which produced accurate results in Table \ref{tab:model_table1}(a), has decreased in predictive performance, with a positive predictive value of approximately $0.5$ for both small and large sample sizes, which indicates that the Support Vector Machine is now producing a large number of false positive predictions. The remaining models again produce AUC values of approximately 0.5, and so we can conclude that these techniques are failing to learn the data, and are producing BVD predictions randomly.  

Table \ref{tab:model_table1}(c) contains modelling results for data with a class imbalance of $95\%$ BVD-negative herds and $5\%$ BVD-positive herds, which was then resampled using SMOTE, to obtain a class balance. Both the Random Forest and the XGBoost model continue to produce accurate predictions for BVD within herds, at both the small- and large sample sizes. The predictive performance of the Support Vector Machine continues to decrease, suggesting that the Support Vector Machine operates most efficiently when classes are balanced, and producing artificially balanced classes via resampling does not sufficiently compensate for the underlying class imbalance. 

Table \ref{tab:model_table1}(d) contains modelling results for data with a class imbalance of $99\%$ BVD-negative herds and $1\%$ BVD-positive herds, which was resampled using SMOTE, to obtain a class balance. At this level of class imbalance, at the small sample size, none of the examined modelling techniques are capable of producing accurate BVD predictions. According to the AUC values, the Random Forest is distinguishing best between positive and negative classes. However, the positive predictive values and sensitivity for this model suggest that it is producing a large number of false positive and false negative predictions. At the large sample size, the Random Forest continues to produce accurate BVD predictions, while the XGBoost model also displays an ability to avoid false negatives through its high sensitivity value.

The data in Table \ref{tab:model_table1} has been resampled using SMOTE to address the class imbalance. However, it can be seen that as the class imbalance becomes more extreme, the effectiveness of SMOTE appears to wane. There are a few possible reasons for this. When the class imbalance is extreme, the minority class is severely under-represented. SMOTE uses the existing minority class samples to produce new samples, and so with few samples to work with, the generated data may be less representative of the underlying distribution of the minority class. This can lead to overfitting to the minority class and a model with poor ability to generalise. In cases of class imbalance, the machine learning method may struggle to correctly classify the majority class after the data has been resampled using SMOTE. As SMOTE increases the representation of the minority class, it can lead to a higher number of false positive predictions (represented in Table \ref{tab:model_table1} as a decrease in positive predictive value with increasing class imbalance).

In Table \ref{tab:model_table2}, we examine the predictive performance of these same machine learning techniques, using data that has not been resampled using SMOTE.

In Table \ref{tab:model_table2}(a), we assess the predictive accuracy of these machine learning techniques on data with $90\%$ BVD-negative herds, and $10\%$ BVD-positive herds. The Random Forest again displays the greatest predictive performance, and seems unaffected by the data imbalance. All other modelling techniques display a marked decrease in predictive performance. For large sample sizes, the Generalised Linear Models, LASSO regression, ridge regression and elastic net regression predict all observations in the dataset as belonging to the majority class, and so there are no true positive predictions with which to calculate positive predictive values, sensitivity or the F1-score. 

Results in Table \ref{tab:model_table2}(b) (using data with $95\%$ BVD-negative herds, and $5\%$ BVD-positive herds)  and Table \ref{tab:model_table2}(c) (using data with $99\%$ BVD-negative herds, and $1\%$ BVD-positive herds) are similar to those in Table \ref{tab:model_table2}(a). Predictive performance of all modelling techniques except the Random Forest decrease with increasing class imbalance. At both the small and large sample size, the Random Forest maintains its predictive performance at all levels of class imbalance assessed. When compared to the decrease in predictive performance for the Random Forest in Table \ref{tab:model_table1}(d), it appears that resampling the data using SMOTE adversely affected model accuracy, as the model may have over-fitted to the synthetic samples, which may not fully represent the complexity of the minority class.

\begin{table}[H]
    \centering
    \begin{tabular}{ c| c c c c | c c c c }

    \hline
     \multicolumn{9}{c}{(a) $10\%$ Positive Herds}\\
     \hline
     & \multicolumn{4}{|c}{Small Sample Size}& \multicolumn{4}{|c}{Large Sample Size}\\
    Model  & PPV & Sensitivity & F1 & AUC   & PPV & Sensitivity & F1 & AUC\\ 
    \hline
    GLM & 0.093 &0.087 &0.091&0.506 &0.000 &0.000 &- &0.523\\
    Random Forest & 0.987 &\textbf{1.000} &0.993&\textbf{0.999}&1.000 &\textbf{0.998} &0.998 &\textbf{1.000}\\
      LASSO & 0.203 & 0.150&0.172&0.516& 0.000&0.000 &- &0.500\\
    Ridge & 0.333 & 0.012&0.0240&0.529& 0.000&0.000 &- &0.500\\
    Elastic Net& 0.203 & 0.150&0.172&0.516& 0.000&0.000 &- &0.500\\
    SVM &0.681&0.400&0.509&0.910&0.765&0.600&0.672&0.972\\
    XGBoost &0.353 &1.000&0.522&0.999&0.356&0.998&0.524&0.992\\
    \hline

    \multicolumn{9}{c}{(b) $5\%$ Positive Herds}\\
    \hline
     & \multicolumn{4}{|c}{Small Sample Size}& \multicolumn{4}{|c}{Large Sample Size}\\
    Model  & PPV & Sensitivity & F1 & AUC   & PPV & Sensitivity & F1 & AUC\\ 
    \hline
    GLM & 0.032  & 0.050&0.039&0.641&0.000 &0.000 &- &0.535\\
    Random Forest &1.000   &\textbf{1.000} &1.000&\textbf{1.000}&0.986 &\textbf{1.000} &0.993 &\textbf{0.999}\\
    LASSO & 0.285 & 0.050&0.085&0.580& 0.000&0.000 &- &0.500\\
    Ridge & 0.285 & 0.050&0.085&0.595& 0.000&0.000 &- &0.500\\
    Elastic Net&0.286  & 0.051&0.086&0.596& 0.000&0.000 &- &0.500\\
    SVM &0.642&0.225&0.333&0.938&0.639&0.440&0.521&0.969 \\
    XGBoost &0.206&1.000&0.341&0.998&0.210&0.998&0.342&0.999\\
    \hline

    \multicolumn{9}{c}{(c) $1\%$ Positive Herds}\\
    \hline
     & \multicolumn{4}{|c}{Small Sample Size}& \multicolumn{4}{|c}{Large Sample Size}\\
    Model  & PPV & Sensitivity & F1 & AUC   & PPV & Sensitivity & F1 & AUC\\ 
    \hline
    GLM & 0.032 & 0.125&0.051&0.485&0.000 &0.000 &- &0.539\\
    Random Forest& 1.000 & 0.750 &0.857&\textbf{0.996}& 0.986&\textbf{1.000} &0.993 &\textbf{0.999}\\
    LASSO & 0.250 & 0.125&0.166&0.682& 0.000&0.000 &- &0.500\\
    Ridge & 0.000 &0.000 &-&0.391& 0.000&0.000 &- &0.500\\
    Elastic Net &0.000&0.000&-&0.500&0.000&0.000&-&0.500\\
    SVM &0.500  & 0.125&0.200&0.597& 0.000&0.000 &- &0.500\\
    XGBoost &0.047 &\textbf{1.000}&0.091&0.996&0.052&0.998&0.099&0.993\\
    \hline
    \end{tabular}
    \caption{Classifier results for  (a) $10\%$ positive herds, (b) $5\%$ positive herds, and (c) $1\%$ positive herds, using data that has not been resampled with SMOTE.In each case, the model with the highest AUC, and the model with the highest sensitivity are highlighted.}
    \label{tab:model_table2}
\end{table}

In Table \ref{tab:model_table3} we present the results of several anomaly detection algorithms. Because these techniques are intended to identify anomalous observations or outliers, we might expect that the predictive performance of these models will improve as the class imbalance becomes more pronounced. In Table \ref{tab:model_table3}(a), the anomaly detection algorithms are trained on data with balanced classes. At both small and large sample sizes, none of the algorithms perform particularly well, with an AUC of approximately $0.5-0.7$ for all algorithms examined. Results are not provided for the angle-based outlier detection algorithm at a large sample size due to issues with computational complexity at large scales. 

In Table \ref{tab:model_table3}(b), we present the results of anomaly detection algorithms trained on data where $90\%$ of herds are BVD-negative, and $10\%$ are BVD-positive. At both small and large sample sizes, the predictive accuracy (as determined by the AUC and the sensitivity) of the local outlier factor method, isolation forests, angle-based outlier detector and k-nearest neighbours has increased with the worsening class imbalance. However, the positive predictive value associated with all algorithms has decreased, which suggests that there is an increased number of false positive predictions. The results associated with the Mahalanobis distance classifier and the autoencoder are similar to those in Table \ref{tab:model_table3}, suggesting that for these algorithms, the increasing data imbalance has not improved the ability to identify anomalies.

Table \ref{tab:model_table3}(c) and Table \ref{tab:model_table3}(d) contain the results of the anomaly detection algorithms trained on data with $5\%$ and $1\%$ BVD-positive herds, respectively. As the class imbalance increases, the positive predictive value of all algorithms decreases, which indicates that all algorithms are producing large numbers of false positive predictions.  In terms of diagnostic test sensitivity, the isolation forest has the greatest predictive performance, followed by the local outlier factor model. This tells us that these algorithms do not suffer from issues associated with false negative predictions. 

The positive predictive value is notably low for most methods, reflecting the difficulty in identifying true anomalies when the dataset is highly imbalanced. The F1-score, which balances positive predictive value and sensitivity, is quite low across all methods. Even with high sensitivity, the low positive predictive value results in reduced F1-scores. For instance, the isolation forest, with its high sensitivity, achieves F1-scores of 0.097 (small sample) and 0.120 (large sample). This reflects the difficulties in achieving a balance between false positives and true positives in extremely imbalanced data.

The combination of high sensitivity and low positive predictive value reflects an important trade-off that often occurs in imbalanced classification tasks, especially in anomaly detection where the minority class is extremely rare. 

The AUC metric, which measures the ability of the model to distinguish between classes, shows a much clearer distinction between models. The isolation forest achieves the highest AUC at 0.982 for the small sample and 0.962 for the large sample, making it the best performer in terms of ranking anomalies relative to normal instances. The local outlier factor also performs reasonably well with AUC values of 0.928 and 0.850, suggesting that, while its positive predictive value and sensitivity are lower, it is still capable of ranking observations effectively by their anomaly scores. 

Given the high imbalance in the dataset, the isolation forest and local outlier factor detection algorithm stand out as the more robust methods in terms of both detection capability (sensitivity) and discrimination (AUC). IF particularly demonstrates strong performance, maintaining high sensitivity and AUC even with the extreme imbalance. In contrast, models like the autoencoder, angle-based outlier detector, and k-nearest neighbours struggle significantly under this imbalance, as evidenced by their low F1-scores and AUC values.

When sensitivity is high but positive predictive value is low, this implies that the model is correctly identifying many of the true anomalies, but at the same time, it is also incorrectly classifying many normal observations as anomalies. There are several possible reasons for this. In a dataset where $99\%$ of observations belong to the majority class, the model is presented with far more opportunities to incorrectly label a normal observation as an anomaly. As only $1\%$ of observations represent the positive class ($8$ out of $800$ observations at the small scale and $100$ out of $10,000$ observations at the large scale), any false positive prediction drastically reduces the positive predictive value, as there are so few true positive cases to begin with. Even a model that is relatively good at detecting anomalies (reflected by a high sensitivity) can have its positive predictive value dragged down by a number of incorrect positive predictions. For example, in Table \ref{tab:model_table3}(d), at the large sample size the isolation forest has a sensitivity of 0.960 which indicates that $96\%$ of the actual anomalies were successfully detected. However, with a positive predictive value of $0.064\%$, only $6.4\%$ of observations predicted as anomalous were correct. The remaining $93.6\%$ of flagged observations actually belonged to the majority class, highlighting a high false positive rate. 

In many real-world anomaly detection applications, high sensitivity is often more critical than high positive predictive value. In the case of BVD prediction, missing a positive case of BVD (false negative) may have severe consequences, so it's often acceptable to tolerate a higher number of false positives, if it means that most of the true positives are caught. However, the cost of false positives is also a factor. In cases where investigating each flagged anomaly is resource-intensive, low precision can be problematic. 

These models were trained to maximise the F1-score - they were trained to find an optimal balance to maximise both the positive predictive value and the sensitivity. However, highly imbalanced datasets can favour sensitivity - in a dataset where only $1\%$ of observations are anomalies, the classifier may make a lot of false positive predictions to increase sensitivity. This occurs because it is easier for the model to cast a wider net and capture more anomalies (increasing sensitivity), even though many of the predictions are incorrect. 

This behaviour is common in anomaly detection tasks where missing an anomaly (false negative) is considered more costly than flagging normal instances as anomalies (false positives). In such cases, the F1-score optimisation tends to lean towards higher sensitivity at the expense of positive predictive value.

 \begin{table}[H]
    \centering
    \begin{tabular}{ c| c c c c | c c c c }
    \hline 
    \multicolumn{9}{c}{(a) $50\%$ Positive Herds}\\
    \hline
     & \multicolumn{4}{|c}{Small Sample Size}& \multicolumn{4}{|c}{Large Sample Size}\\
    Model  & PPV & Sensitivity & F1 & AUC   & PPV & Sensitivity & F1 & AUC\\ 
    \hline
    LOF & 0.766 & 0.230 &0.353&0.734& 0.653 & 0.195 & 0.300 & 0.521\\
    IF &0.716&0.215&0.331&0.669&0.671&0.201&0.309&\textbf{0.638} \\
    ABOD &0.461&\textbf{0.830}&0.592&\textbf{0.768}&-&-&-&- \\
    KNN &0.650&0.260&0.371&0.609&0.547&0.219&0.312&0.519\\
    Mahalanobis &0.675&0.270&0.385&0.651&0.657&\textbf{0.263}&0.375&0.634\\
    Autoencoder &0.587&0.235&0.335&0.535&0.582&0.232&0.332&0.532\\
    
    \hline
     \multicolumn{9}{c}{(b) $10\%$ Positive Herds}\\
     \hline
     & \multicolumn{4}{|c}{Small Sample Size}& \multicolumn{4}{|c}{Large Sample Size}\\
    Model  & PPV & Sensitivity & F1 & AUC   & PPV & Sensitivity & F1 & AUC\\ 
    \hline
    LOF & 0.441 &\textbf{0.663} &0.530&0.897 & 0.354 & \textbf{0.532} & 0.425 & 0.847\\
    IF &0.383&0.575&0.460&\textbf{0.898}&0.345&0.518&0.414&\textbf{0.889}\\
    ABOD &0.073&0.662&0.133&0.809&-&-&-&-\\
    KNN &0.250&0.500&0.330&0.726&0.162&0.324&0.216&0.597\\
    Mahalanobis &0.118&0.237&0.158&0.591&0.012&0.243&0.162&0.596\\  
    Autoencoder &0.112&0.225&0.150&0.513&0.127&0.255&0.170&0.531\\
    \hline

    \multicolumn{9}{c}{(c) $5\%$ Positive Herds}\\
    \hline
     & \multicolumn{4}{|c}{Small Sample Size}& \multicolumn{4}{|c}{Large Sample Size}\\
    Model  & PPV & Sensitivity & F1 & AUC   & PPV & Sensitivity & F1 & AUC\\ 
    \hline
    LOF & 0.258  & 0.775&0.387&0.912& 0.224 & 0.674& 0.337 & 0.886\\
    IF &0.292&\textbf{0.875}&0.437&\textbf{0.944}&0.281&\textbf{0.842}&0.421&\textbf{0.934}\\
    ABOD &0.032&0.570&0.061&0.795&-&-&-&-\\
    KNN &0.125&0.500&0.200&0.730&0.092&0.366&0.146&0.632\\
    Mahalanobis &0.068&0.275&0.110&0.612&0.006&0.258&0.103&0.615\\ 
    Autoencoder &0.037&0.150&0.060&0.473&0.060&0.240&0.096&0.521\\
    \hline

    \multicolumn{9}{c}{(d) $1\%$ Positive Herds}\\
    \hline
     & \multicolumn{4}{|c}{Small Sample Size}& \multicolumn{4}{|c}{Large Sample Size}\\
    Model  & PPV & Sensitivity & F1 & AUC   & PPV & Sensitivity & F1 & AUC\\ 
    \hline
    LOF & 0.058 & 0.875&0.109&0.928& 0.056& 0.850& 0.106&0.919\\
    IF &0.066&\textbf{0.997}&0.125&\textbf{0.982}&0.064&\textbf{0.960}&0.120&\textbf{0.962}\\
    ABOD &0.005&0.500&0.010&0.835&-&-&-&-\\
    KNN &0.032&0.625&0.059&0.774&0.023&0.470&0.044&0.682\\
    Mahalanobis &0.031&0.625&0.059&0.723&0.016&0.320&0.031&0.653\\
    Autoencoder &0.016&0.250&0.031&0.688&0.012&0.250&0.023&0.525\\
    \hline
    \end{tabular}
    \caption{Classifier results for (a) $50\%$ positive herds, (b) $10\%$ positive herds, (c) $5\%$ positive herds, and (d) $1\%$ positive herds, using data that has not been resampled with SMOTE. In each case, the model with the highest AUC, and the model with the highest sensitivity are highlighted.}
    \label{tab:model_table3}
\end{table}

In Table \ref{tab:model_table4} we present the results of weighted classification methods, as described in Section \ref{imbalanced_data}. Similarly to the tables shown above, Table \ref{tab:model_table4} is divided into four sections, depending on the percentage of herds that are simulated positive for BVD. 

Here we examine generalised linear models, random forests, regularised regression, support vector machines and XGBoost. All were examined at small and large sample sizes, except support vector machines, which were only examined at a small sample size because of prohibitive model runtimes at a large sample size. In each case, the model with the highest sensitivity and AUC is highlighted. In cases where two models perform at comparable levels, both are highlighted. 

In Table \ref{tab:model_table4}(a) we examine the scenario where positive and negative herds are balanced. At both small- and large sample sizes, the random forest and XGBoost models are tied for best performance in terms of both sensitivity and AUC value. In particular, the XGBoost model provides near-perfect predictions, with positivie predictive value and sensitivity value of $1.00$, indicating that there are no false positive or false negative predictions occurring. 

This pattern repeats for Table \ref{tab:model_table4}(b) to Table \ref{tab:model_table4}(d), as the random forest and XGBoost model compete for highest predictive power, while the generalised linear model, regularised regression and support vector machines perform poorly. The predictive power of both the random forest and XGBoost model does fall slightly as class imbalance becomes more severe in Table \ref{tab:model_table4}(d). However, with sensitivity values of between $0.7$ and $0.9$, and AUC values of approximately $0.9$, these models are still performing reliably well despite class imbalance.

\begin{table}[H]
    \centering
    \begin{tabular}{ c| c c c c | c c c c }
    \hline 
    \multicolumn{9}{c}{(a) $50\%$ Positive Herds}\\
    \hline
     & \multicolumn{4}{|c}{Small Sample Size}& \multicolumn{4}{|c}{Large Sample Size}\\
    Model  & PPV & Sensitivity & F1 & AUC   & PPV & Sensitivity & F1 & AUC\\ 
    \hline
    GLM & 0.462 & 0.479 &0.471&0.509 &0.494 &0.501 &0.497 &0.504\\
    Random Forest&1.000 & 0.997&0.998&\textbf{1.000}&0.999 &\textbf{0.998}&0.998 &\textbf{1.000}\\
    LASSO & 0.562 & 0.521&0.541&0.521& 0.522&0.493 &0.507 &0.509\\
    Ridge & 0.505 & 0.489&0.496&0.511& 0.523&0.495 &0.509 &0.505\\
    Elastic Net& 0.503 & 0.489&0.495&0.510& 0.576&0.502 &0.536 &0.504\\
    SVM &0.860&0.924&0.891&0.963&-&-&-&-\\
    XGBoost &1.000&\textbf{1.000}&1.000&\textbf{1.000}&1.000&\textbf{0.998}&0.999&\textbf{1.000}\\
 
    \hline
     \multicolumn{9}{c}{(b) $10\%$ Positive Herds}\\
     \hline
     & \multicolumn{4}{|c}{Small Sample Size}& \multicolumn{4}{|c}{Large Sample Size}\\
    Model  & PPV & Sensitivity & F1 & AUC   & PPV & Sensitivity & F1 & AUC\\ 
    \hline
    GLM & 0.287 & 0.100&0.148&0.503&0.450 &0.101 &0.165 &0.491\\
    Random Forest&1.000& 0.963&0.981&\textbf{0.999}&0.999 &\textbf{1.000} &0.999 &\textbf{1.000}\\
    LASSO &0.325  &0.082 &0.131&0.546& 0.562&0.102 &0.173 &0.506\\
    Ridge & 0.350 &0.084 &0.136&0.556& 0.514&0.099 &0.166 &0.500\\
    Elastic Net& 0.375 & 0.075&0.125&0.601& 0.729&0.101 &0.177 &0.505\\
    SVM &0.287&0.851&0.429&0.924&-&-&-&-\\
    XGBoost &1.000& \textbf{0.987}&0.984&\textbf{0.999}&1.000&\textbf{1.000}&1.000&\textbf{1.000}\\
    \hline

    \multicolumn{9}{c}{(c) $5\%$ Positive Herds}\\
    \hline
     & \multicolumn{4}{|c}{Small Sample Size}& \multicolumn{4}{|c}{Large Sample Size}\\
    Model  & PPV & Sensitivity & F1 & AUC   & PPV & Sensitivity & F1 & AUC\\ 
    \hline
    GLM & 0.100  & 0.031&0.047&0.593&0.482 &0.056 &0.101 &0.533\\
    Random Forest&1.000& \textbf{1.000}&1.000&\textbf{1.000}&1.000 &\textbf{0.986} &0.993 &\textbf{0.999}\\
    LASSO & 0.800 & 0.050&0.094&0.500& 0.582&0.053 &0.097 &0.524\\
    Ridge & 0.375 & 0.053&0.093&0.537& 0.578&0.057 &0.105 &0.551\\
    Elastic Net& 0.800 & 0.051&0.095&0.500& 0.582&0.050 &0.095 &0.530\\
    SVM &0.000&0.000&-&0.502&-&-&-&-\\
    XGBoost &1.000&\textbf{1.000}&1.000&\textbf{1.000}&1.000&0.984&0.992&\textbf{0.999}\\
    \hline

    \multicolumn{9}{c}{(d) $1\%$ Positive Herds}\\
    \hline
     & \multicolumn{4}{|c}{Small Sample Size}& \multicolumn{4}{|c}{Large Sample Size}\\
    Model  & PPV & Sensitivity & F1 & AUC   & PPV & Sensitivity & F1 & AUC\\ 
    \hline
    GLM &0.125  &0.033 &0.052&0.495&0.320 &0.011 &0.022 &0.529\\
    Random Forest&1.000&0.727&0.842&\textbf{0.999}& 0.970 &\textbf{0.970} &0.970 &\textbf{0.999}\\
    LASSO & 0.375 & 0.011&0.021&0.516& 0.930&0.010 &0.020 &0.524\\
    Ridge & 0.125 & 0.004&0.009&0.524& 0.440&0.011 &0.023 &0.520\\
    Elastic Net& 0.250 & 0.006&0.012&0.574& 0.993&0.011 &0.021 &0.525\\
    SVM &0.000&0.000&-&0.512&-&-&-&-\\
    XGBoost &0.875&\textbf{0.875}&0.875&0.985&0.990&0.733&0.842&\textbf{0.999}\\
    \hline
    \end{tabular}
    \caption{Weighted classifier results for (a) $50\%$ positive herds, (b) $10\%$ positive herds, (c) $5\%$ positive herds, and (d) $1\%$ positive herds, using data that has not been resampled with SMOTE. In each case, the model with the highest AUC, and the model with the highest sensitivity are highlighted.}
    \label{tab:model_table4}
\end{table}

Of the methodologies examined in this simulation study, the ones that showed the most promise in terms of predicting BVD in imbalanced datasets were applied to predicting BVD in Irish cattle herds. When using imbalanced data that has not been resampled (Table \ref{tab:model_table1}, the methods that performed best were the random forest and XGBoost methods. When using data and "correcting" the imbalance via re-sampling (Table \ref{tab:model_table2}), the models with the best predictive power were the random forest, XGBoost, elastic net and LASSO models. Of the anomaly detectors (Table \ref{tab:model_table3}), the best performers were the local outlier factor and isolation forests, and of the weighted classifiers (Table \ref{tab:model_table4}), the best-performing models were the random forest and XGBoost methods.

\section{Irish Cattle Herds}
\label{data}
The dataset used in this study comprises information on cattle herds in the Republic of Ireland from 2013 to 2023, a critical period in the national effort to eradicate BVD. The primary objective of this analysis was to apply machine learning methods, as outlined in Section \ref{methods}, to predict BVD occurrence, thereby enabling a more targeted approach to disease monitoring and testing.

As a result of the success of the eradication programme in significantly reducing BVD prevalence, the dataset exhibits substantial class imbalance, with the vast majority of herds testing negative for BVD. Table \ref{tab:cases_prop} presents the annual proportion of BVD-positive and BVD-negative herds, illustrating the steady decline in positive cases over time. This class imbalance poses challenges for training and evaluating predictive models. To address these challenges, several strategies were employed, including resampling techniques, tailored evaluation metrics, and anomaly detection methods that treated BVD-positive cases as outliers.

\begin{table}[H]
    \centering
    \begin{tabular}{c c c}
    \hline
        Year &  Positive BVD Cases ($\%$) & Negative BVD Cases ($\%$)\\
        \hline
        2014 & 6.82 & 93.18\\
        2015 & 5.27 & 94.73\\
        2016 & 2.82& 97.18\\
        2017 & 1.64& 98.36\\   
        2018 & 0.86& 99.14\\
        2019 & 0.61& 99.39\\
        2020 & 0.41& 99.59\\
        2021 & 0.37 & 99.63\\
        2022 & 0.31 &99.69\\
        2023 &0.27&99.73\\
        \hline   
    \end{tabular}
    \caption{Annual percentage of BVD-positive and BVD-negative herds, where a BVD-positive herd experiences at least one case of BVD, and a BVD-negative herd does not experience any cases of BVD.}
    \label{tab:cases_prop}
\end{table}

Given these challenges, only models with the highest predictive performance in preliminary simulation studies were applied to the full dataset. These models included:
\begin{itemize}
    \item Random forests and XGBoost, both with and without resampling and class-weight adjustments.
    \item Elastic net and LASSO regression, applied to resampled data.
    \item The isolation forest anomaly detection algorithm, which identified BVD-positive cases as outliers.
\end{itemize}

The dataset contains several covariates, including:
\begin{itemize}
    \item Binary indicators for each herd's BVD status (1 for BVD-positive, 0 for BVD-negative) from 2013 to 2023.
    \item The number of calves born annually per herd.
    \item The number of non-dairy animals in each herd per year.
    \item Animal movement data (the number of individuals that are sent to a factory, a knackery, a farm or a mart, and the number of exports).
    \item The number of stillborn animals.
    \item The degree of the herd. Degree measures how many neighbours a herd has. 
    \item The betweenness of the herd. Betweenness measures how often a herd acts as a bridge along the shortest path between other herds. High betweenness herds may play a role in connecting distant parts of the network and can facilitate the spread of BVD between otherwise unconnected herds.
    \item The closeness of the herd. Closeness measures how central a herd is within the network of Irish cattle herds. A herd with high closeness has shorter average paths to all other herds, meaning it is well-connected and has a relatively high likelihood of being exposed to BVD in an outbreak.
    \item Local disease density, calculated for each herd as the number of BVD-positive neighbours divided by the total number of neighbours for the previous year.
\end{itemize}

These covariates provide critical information for predicting BVD occurrence, supporting the development of more targeted surveillance and intervention strategies. Models were run to predict BVD status of herds in 2023, using covariate data from 2021 and 2022. All models were run including all two-way interaction terms for the covariates, which resulted in a model with $230$ predictors and $93,330$ herds, of which $250$ were positive in 2023. K-fold cross-validation with $k=5$ was performed, which involved splitting the data into five subsets, training each method on four of the subsets, and then testing predictions on the final subset. This process was repeated five times, with each subset serving as the test set once while the remaining four subsets formed the training set. The results from all iterations were then combined. This approach helps mitigate overfitting and ensures that the model generalises well to unseen data.

\begin{table}[H]
    \centering
    \begin{tabular}{lcccccc}
    \hline
       \multicolumn{7}{c}{Imbalanced Classification}\\
        Model  & Sensitivity & AUC &TP & FP & TN & FN\\   
        \hline
        Random Forest  & 0.876 & 0.536& 219 & 46718 & 46362 & 31\\
        XGBoost  &0.364 & 0.497 &91& 37601& 55479 &159\\

    \hline
       \multicolumn{7}{c}{Resampled Classification}\\
        Model  & Sensitivity & AUC &TP & FP & TN & FN\\   
        \hline

        Random Forest & 0.724 & 0.485 & 181 & 46784 & 46296 &69\\  
        XGBoost  &0.756 & 0.506 &189&46779&46301&61\\ 
        Elastic Net  &0.868 & 0.507 &217&46742&46338&33\\
        LASSO  &0.384 & 0.501 &96&37598&55482&154 \\ 
    \hline
       \multicolumn{7}{c}{Weighted Classification}\\
        Model  & Sensitivity & AUC & TP & FP & TN & FN\\   
        \hline

        Random Forest  & 0.844 & 0.526 & 211 & 46129 & 46951 & 39\\
        XGBoost  &0.624 & 0.499 &156&39291&53789&94\\
        \hline
       \multicolumn{7}{c}{Anomaly Detectors}\\
        Model  & Sensitivity & AUC &TP & FP & TN & FN\\   
        \hline

        Isolation Forest &0.396 &0.697&99&36629&56451&151\\

        \hline
    \end{tabular}
    \caption{Comparison of true positives, false positives, true negatives, and false negatives across ML models.}
    \label{tab:result_tab}
\end{table}

Table \ref{tab:result_tab} presents a comparative analysis of various machine learning models for predicting 2023 infection status, evaluated across multiple classification strategies: imbalanced classification, resampled classification, weighted classification, and anomaly detection. The primary objective of the analysis is to maximise the number of true positive predictions (TP) while balancing false positive predictions (FP). Given the highly imbalanced dataset, where only $250$ of the $93,330$ herds are truly positive for BVD, models are evaluated based on sensitivity, area under the curve (AUC), and their ability to correctly identify positive cases.

The Random Forest and XGBoost models were applied directly to the imbalanced dataset. Among the imbalanced classification models, the Random Forest model achieved the highest sensitivity (0.876), correctly identifying 219 out of the 250 truly positive cases, with only 31 false negatives (FN). However, this came at the cost of a large number of false positives (46,718), meaning a significant proportion of herds were incorrectly classified as positive. In contrast, XGBoost had a much lower sensitivity (0.364), identifying only 91 true positives, but with a substantial reduction in false positives (37,601).

Resampling techniques were employed to balance the dataset. When resampling techniques were applied, Elastic Net outperformed other models in terms of sensitivity (0.868) while slightly reducing false positives (46,742). The resampled XGBoost model also improved over its imbalanced counterpart, achieving a sensitivity of 0.756 with 189 correctly identified positive cases. The LASSO model, however, performed poorly in terms of sensitivity (0.384), identifying only 96 positive cases while producing 37,598 false positive predictions.

Weighting techniques were used to account for the class imbalance. Weighted classification approaches led to a more balanced trade-off between sensitivity and false positives. The weighted Random Forest model achieved a sensitivity of 0.844, correctly identifying 211 positive cases with a slightly reduced number of false positives (46,129) compared to the unweighted counterpart. Weighted XGBoost had lower sensitivity (0.624) but further decreased false positives (39,291).

Of the anomaly detection algorithms examined as part of the simulation study, only the isolation forest was chosen to examine the full dataset. Isolation forest performed poorly in terms of sensitivity (0.396), identifying only 99 positive cases, despite having a relatively lower false positive count (36,629).

Given the objective of maximising true positives while maintaining a reasonable balance with false positives, models such as the random forest (imbalanced and weighted) and Elastic Net (resampled) appear to be the most effective. However, adjusting the classification threshold could further improve sensitivity at the expense of increasing false positives. Lowering the threshold would capture more true positives but may also introduce more false positives, which would need to be managed depending on the practical implications of misclassification in this context.

Overall, different modelling approaches lead to varying trade-offs between false positives and false negatives. In an imbalanced setting, prioritising sensitivity often increases false positive levels. The non-resampled random forest was ultimately chosen as the model that offered the best balance between capturing positive BVD cases and minimising false positive predictions. This model correctly identifies $219$ of the total $250$ positive BVD cases in 2023. Associated with this are $46,718$ false positive predictions. While high, this reflects a $50\%$ reduction in unnecessary testing, as the current strategy involves blanket-testing of all herds. By reducing the classification threshold, it is possible to further increase the number of positive cases identified at the cost of a large increase in false positive predictions. For example, if the classification threshold is reduced from $0.5$ to $0.1$, the sensitivity increases to $0.98,$ correctly identifying $245$ of the $250$ positive herds but the false positive predictions increases to $83,799$, and so the decrease in the number of herds requiring tests is now only $10\%$,which could place undue testing pressure on healthy herds.

Resampling and weighted classification strategies offered improved sensitivity performance, but with increased false positive rates. Traditional machine learning models, particularly Random Forest and XGBoost, demonstrated competitive performance. These results highlight the importance of selecting models based on the specific objectives of disease prediction, particularly in scenarios where minimising false negatives is crucial to preventing undetected outbreaks.

 This model supports the primary goal of minimising missed infections while maintaining a practical testing burden for BVD eradication in Ireland. Future work could explore further refinements, such as hybrid approaches combining classification and anomaly detection methods or adjusting class weights dynamically to handle evolving case distributions in BVD surveillance.

\section{Discussion}
\label{discussion}
The results of this study highlight the challenges and trade-offs associated with predicting BVD occurrences in Irish cattle herds using machine learning models. The significant class imbalance in the data prompted the use of various strategies, including resampling techniques, weighted classification and anomaly detection. Each approach had distinct advantages and limitations, which influenced model selection. 

Random forest models consistently demonstrated high sensitivity, making them effective at identifying positive BVD cases. The random forest, implemented on the imbalanced data, correctly classified $219$ out of $250$ positive cases, achieving the highest sensitivity ($0.876$). However, this came at the cost of a substantial number of false positive prediction ($46,718$) which could lead to unnecessary testing. Weighted classification slightly reduced false positive predictions ($46,129$), while maintaining a high sensitivity ($0.844$), making it a viable alternative. 

Among resampled classifiers, Elastic Net provided strong predictive performance, achieving a sensitivity of $0.868$ while slightly lowering false positive predictions compared to the imbalanced random forest. Resampling effectively balanced sensitivity and specificity, but with trade-offs in false positive counts. XGBoost models, while generally performing well in machine learning applications, struggled in this imbalanced setting. Even when resampling and weighting were applied, their sensitivity remained moderate compared to the random forest and elastic net. 

Anomaly detection, represented by the Isolation Forest, performed poorly in this context. Despite a reasonable false positive count, its low sensitivity ($0.396$) rendered it ineffective for practical use. This suggests that BVD-positive herds do not exhibit sufficiently distinct patterns to be consistently identified as anomalies. 

Given the goal of maximising true positive predictions while maintaining a manageable levels of false positive predictions, the imbalanced random forest model was ultimately chosen. This model correctly identified $219$ positive cases while reducing unnecessary tests by approximately $50\%$, compared to the blanket-testing approach currently in place. However, further reductions in the classification threshold could improve sensitivity at the cost of increasing false positive predictions, which would place additional burdens on unaffected herds.

These findings have significant implications for disease surveillance and control. The ability to target high-risk herds for testing could improve resource allocation and accelerate BVD eradication efforts. However, balancing sensitivity and false positive predictions remains a key challenge. Future work could explore dynamic weighting strategies to adapt to changing prevalence rates and ensemble methods that combine the predictions of individual machine learning methods to improve predictive power. 

Overall, this study demonstrates the potential of machine learning for targeted disease surveillance, while also emphasising the importance of selecting models that align with specific policy and practical constraints in animal health management.

\bibliographystyle{apalike}  
\bibliography{references}

\begin{thebibliography}{}

\bibitem[{Animal Health Ireland}, 2025]{ahi2024}
{Animal Health Ireland} (2025).
\newblock Bvd programme results.
\newblock Accessed: 2024-10-20.

\bibitem[Breiman, 2001]{breiman2001random}
Breiman, L. (2001).
\newblock Random forests.
\newblock {\em Machine learning}, 45:5--32.

\bibitem[Breiman et~al., 1984]{breiman1984classification}
Breiman, L., Friedman, J., Stone, C.~J., and Olshen, R.~A. (1984).
\newblock {\em Classification and regression trees}.
\newblock CRC press.

\bibitem[Breunig et~al., 2000]{breunig2000lof}
Breunig, M.~M., Kriegel, H.-P., Ng, R.~T., and Sander, J. (2000).
\newblock Lof: identifying density-based local outliers.
\newblock In {\em Proceedings of the 2000 ACM SIGMOD international conference on Management of data}, pages 93--104.

\bibitem[Chawla et~al., 2002]{chawla2002smote}
Chawla, N.~V., Bowyer, K.~W., Hall, L.~O., and Kegelmeyer, W.~P. (2002).
\newblock Smote: synthetic minority over-sampling technique.
\newblock {\em Journal of artificial intelligence research}, 16:321--357.

\bibitem[Chen and Guestrin, 2016]{chen2016xgboost}
Chen, T. and Guestrin, C. (2016).
\newblock Xgboost: A scalable tree boosting system.
\newblock In {\em Proceedings of the 22nd acm sigkdd international conference on knowledge discovery and data mining}, pages 785--794.

\bibitem[Cortes and Vapnik, 1995]{cortes1995support}
Cortes, C. and Vapnik, V. (1995).
\newblock Support-vector networks.
\newblock {\em Machine learning}, 20:273--297.

\bibitem[Golub et~al., 1999]{golub1999tikhonov}
Golub, G.~H., Hansen, P.~C., and O'Leary, D.~P. (1999).
\newblock Tikhonov regularization and total least squares.
\newblock {\em SIAM journal on matrix analysis and applications}, 21(1):185--194.

\bibitem[Hahne et~al., 2008]{hahne2008unsupervised}
Hahne, F., Huber, W., Gentleman, R., Falcon, S., Gentleman, R., and Carey, V. (2008).
\newblock Unsupervised machine learning.
\newblock {\em Bioconductor case studies}, pages 137--157.

\bibitem[Hanley and McNeil, 1982]{hanley1982meaning}
Hanley, J.~A. and McNeil, B.~J. (1982).
\newblock The meaning and use of the area under a receiver operating characteristic (roc) curve.
\newblock {\em Radiology}, 143(1):29--36.

\bibitem[Houe, 1995]{houe1995epidemiology}
Houe, H. (1995).
\newblock Epidemiology of bovine viral diarrhea virus.
\newblock {\em Veterinary Clinics of North America: Food Animal Practice}, 11(3):521--547.

\bibitem[Kriegel et~al., 2008]{kriegel2008angle}
Kriegel, H.-P., Schubert, M., and Zimek, A. (2008).
\newblock Angle-based outlier detection in high-dimensional data.
\newblock In {\em Proceedings of the 14th ACM SIGKDD international conference on Knowledge discovery and data mining}, pages 444--452.

\bibitem[Mahalanobis, 1936]{mahalanobis1936generalised}
Mahalanobis, P.~C. (1936).
\newblock On the generalised distance in statistics.
\newblock {\em Proceedings of the National Institute of Sciences (Calcutta)}, 2:49--55.

\bibitem[Nasteski, 2017]{nasteski2017overview}
Nasteski, V. (2017).
\newblock An overview of the supervised machine learning methods.
\newblock {\em Horizons. b}, 4(51-62):56.

\bibitem[Neethirajan, 2020]{neethirajan2020role}
Neethirajan, S. (2020).
\newblock The role of sensors, big data and machine learning in modern animal farming.
\newblock {\em Sensing and Bio-Sensing Research}, 29:100367.

\bibitem[Nelder and Wedderburn, 1972]{nelder1972generalized}
Nelder, J.~A. and Wedderburn, R.~W. (1972).
\newblock Generalized linear models.
\newblock {\em Journal of the Royal Statistical Society Series A: Statistics in Society}, 135(3):370--384.

\bibitem[{R Core Team}, 2023]{R}
{R Core Team} (2023).
\newblock {\em R: A Language and Environment for Statistical Computing}.
\newblock R Foundation for Statistical Computing, Vienna, Austria.

\bibitem[Rousseeuw and Driessen, 1999]{rousseeuw1999fast}
Rousseeuw, P.~J. and Driessen, K.~V. (1999).
\newblock A fast algorithm for the minimum covariance determinant estimator.
\newblock {\em Technometrics}, 41(3):212--223.

\bibitem[Stott et~al., 2012]{stott2012predicted}
Stott, A.~W., Humphry, R.~W., Gunn, G.~J., Higgins, I., Hennessy, T., O’Flaherty, J., and Graham, D.~A. (2012).
\newblock Predicted costs and benefits of eradicating bvdv from ireland.
\newblock {\em Irish veterinary journal}, 65:1--11.

\bibitem[Tibshirani, 1996]{tibshirani1996regression}
Tibshirani, R. (1996).
\newblock Regression shrinkage and selection via the lasso.
\newblock {\em Journal of the Royal Statistical Society Series B: Statistical Methodology}, 58(1):267--288.

\bibitem[Zhang et~al., 2021]{zhang2021application}
Zhang, S., Su, Q., and Chen, Q. (2021).
\newblock Application of machine learning in animal disease analysis and prediction.
\newblock {\em Current Bioinformatics}, 16(7):972--982.

\bibitem[Zou and Hastie, 2004]{zou2005regularization}
Zou, H. and Hastie, T. (2004).
\newblock Regularization and variable selection via the elastic net.
\newblock {\em Journal of the Royal Statistical Society Series B: Statistical Methodology}, 67(2):301--320.

\end{thebibliography}

\end{document}